\documentclass{article}

\usepackage[colorlinks=true,citecolor=brown,urlcolor=gray,linkcolor=BlueViolet]{hyperref}

\usepackage[accepted]{icml2026}

\usepackage{amsmath}
\usepackage{amssymb}
\usepackage{mathtools}
\usepackage{amsthm}
\usepackage{placeins}
\usepackage{microtype}
\usepackage{graphicx}
\usepackage{subcaption}
\usepackage{booktabs} 

\usepackage{enumitem}
\usepackage{arydshln} 
\usepackage[utf8]{inputenc} 
\usepackage[T1]{fontenc}    

\usepackage{url}            
\usepackage{wrapfig,lipsum,booktabs}       
\usepackage{amsfonts}       
\usepackage{nicefrac}       
\usepackage{xcolor}         
\usepackage[dvipsnames,table]{xcolor}  
\definecolor{lightgreen}{HTML}{dcf5dc} 
\definecolor{lightred}{HTML}{fce4e4}
\usepackage{colortbl}
\usepackage{pifont}
\usepackage{rotating}
\usepackage{makecell}
\usepackage{graphicx}
\usepackage{subcaption} 
\usepackage{bbm}

\usepackage{caption}
\usepackage{subcaption}
\usepackage{xspace}
\usepackage{multirow}

\usepackage{algorithm}

\usepackage[titletoc]{appendix}

\usepackage{pdflscape}

\usepackage{tabularray}

\usepackage{soul}
\usepackage{tcolorbox}
\tcbuselibrary{skins,listings,breakable} %
\newcounter{methodcount}

\usepackage[capitalize,noabbrev]{cleveref}

\usepackage[textsize=tiny]{todonotes}

\icmltitlerunning{\ours: Process Verification of Multi-Agent Systems}

\newcommand{\ours}{\texttt{MAS-ProVe}\xspace}

\newcommand{\blue}[1]{\textcolor{blue}{#1}}

\newlength\savewidth

\newcolumntype{C}{>{\centering\let\newline\\\arraybackslash\hspace{0pt}}m{2cm}}

\usepackage{amsmath,amsfonts,bm,eqnarray}
\usepackage{cancel}
\usepackage{amsthm}
\usepackage{tikz}
\usetikzlibrary{tikzmark}

\newtheorem{innercustomgeneric}{\customgenericname}
\providecommand{\customgenericname}{}
\newcommand{\newcustomtheorem}[2]{%
  \newenvironment{#1}[1]
  {%
   \renewcommand\customgenericname{#2}%
   \renewcommand\theinnercustomgeneric{##1}%
   \innercustomgeneric
  }
  {\endinnercustomgeneric}
}

\newcustomtheorem{customThm}{Theorem}
\newcustomtheorem{customLemma}{Lemma}
\newcustomtheorem{customCor}{Corollary}
\newcustomtheorem{customProposition}{Proposition}

\def\Tabref#1{Table~\ref{#1}}

\def\Figref#1{Fig.~\ref{#1}}

\def\appref#1{Appendix~\ref{#1}}
\def\Secref#1{Sec.~\ref{#1}}

\def\eqref#1{equation~\ref{#1}}

\def\1{\bm{1}}

\DeclareMathAlphabet{\mathsfit}{\encodingdefault}{\sfdefault}{m}{sl}
\SetMathAlphabet{\mathsfit}{bold}{\encodingdefault}{\sfdefault}{bx}{n}

\renewcommand{\frac}{\tfrac}

\definecolor{lightergray}{HTML}{e5e5e5}
\definecolor{lightgray}{HTML}{cccccc}
\sethlcolor{lightergray}
\sethlcolor{lightgray}

\begin{document}

\twocolumn[
  \icmltitle{\ours: Understanding the Process Verification of Multi-Agent Systems}



  \icmlsetsymbol{equaladvise}{$\ddagger$}
  \icmlsetsymbol{lead}{$\dagger$}
  \icmlsetsymbol{intern}{*}

  \begin{icmlauthorlist}
    \icmlauthor{Vishal Venkataramani}{ru}
    \icmlauthor{Haizhou Shi}{intern,ru,sfr}
    \icmlauthor{Zixuan Ke}{lead,sfr}
    \icmlauthor{Austin Xu}{sfr}
    \icmlauthor{Xiaoxiao He}{ru}
    \icmlauthor{Yingbo Zhou}{sfr}
    \icmlauthor{Semih Yavuz}{sfr}
    \icmlauthor{Hao Wang}{equaladvise,ru}
    \icmlauthor{Shafiq Joty}{equaladvise,sfr}
  \end{icmlauthorlist}

  \icmlaffiliation{ru}{Rutgers University}
  \icmlaffiliation{sfr}{Salesforce AI Research} 

  \icmlcorrespondingauthor{Vishal Venkataramani}{vishal.venkataramani@rutgers.edu}
  \icmlcorrespondingauthor{Zixuan Ke}{zixuan.ke@salesforce.com}
  \icmlcorrespondingauthor{Hao Wang}{hw488@cs.rutgers.edu}
  \icmlcorrespondingauthor{Shafiq Joty}{sjoty@salesforce.com}

  \icmlkeywords{Machine Learning, ICML}

  \vskip 0.3in
]



\printAffiliationsAndNotice{
$^*$Work done during an internship at Salesforce AI Research. 
$^\dagger$Project Lead. 
$^\ddagger$Equal Advising.
}

\begin{abstract}
Multi-Agent Systems (MAS) built on Large Language Models (LLMs) often exhibit high variance in their reasoning trajectories. 
{Process verification, which evaluates intermediate steps in trajectories, has shown promise in general reasoning settings, and has been suggested as a potential tool for guiding coordination of MAS; however, its actual effectiveness in MAS remains unclear.}
{To fill this gap, we present \ours, a systematic empirical study of process verification for MAS. Our study spans \emph{three verification paradigms} (LLM-as-a-Judge, reward models, and process reward models), evaluated across \emph{two levels of verification granularity} (agent-level and iteration-level). We further examine \emph{five representative verifiers} and \emph{four context management strategies,} and conduct experiments over \emph{six diverse MAS frameworks} on multiple reasoning benchmarks.}
We find that process-level verification does not consistently improve performance and frequently exhibits high variance, highlighting the difficulty of reliably evaluating partial multi-agent trajectories. Among the methods studied, LLM-as-a-Judge generally outperforms reward-based approaches, with trained judges surpassing general-purpose LLMs. We further observe a small performance gap between LLMs acting as judges and as single agents, and identify a context-length-performance trade-off in verification. Overall, our results suggest that effective and robust process verification for MAS remains an open challenge, requiring further advances beyond current paradigms.
Code is available at \url{https://github.com/Wang-ML-Lab/MAS-ProVe}.
\end{abstract}

\section{Introduction}
\label{sec:intro}
Recent advances in Large Language Models~(LLMs) have enabled the solution of a wide range of tasks requiring careful high-level planning and scheduling~\citep{ke2025surveyfrontiersllmreasoning}, complex logical and mathematical reasoning~\citep{wang2025surveylargelanguagemodels}, and self-reflection for iterative outcome improvement~\citep{shinn2023reflexionlanguageagentsverbal}. 
As these intrinsic capabilities continue to mature, and as the scaling gains of single foundation models begin to plateau, there has been growing interest in constructing coordinated ensembles of LLMs that can collectively understand, decompose, execute, summarize, and reflect on complex tasks~\cite{kim2025sciencescalingagentsystems}. This emerging paradigm of Multi-Agent Systems~(MAS) offers several advantages over Single-Agent Systems~(SAS), including reduced context interference and improved parallelism~\cite{langchain_open_deep_research,Ke2026MASOrchestra}.

Any potential benefits of MAS are accompanied by increased complexity, increasing the number of potential failure points: final outcome correctness in MAS now depends inherently on the intermediate outputs of a variety of coordinated sub-agents. To weed out mistakes before they influence downstream correctness, many MAS frameworks~\citep{shinn2023reflexionlanguageagentsverbal,du2024improving,liang2023encouraging} explicitly incorporate automatic verification sub-agents to check or refine intermediate outputs, while others work incorporate verifiers for further do test-time scaling of MAS~\citep{brown2024largelanguagemonkeysscaling,jin2025two}. Despite the explicit inclusion of verification modules, subpar evaluation remains a bottleneck in the efficacy of MAS~\citep{cemri2025multiagentllmsystemsfail}. This work attempts to understand the fundamental role of automatic evaluation in MAS by systematically studying evaluators, MAS frameworks, and domains, guided by a central research question:
\begin{center}
\textit{Do multi-agent systems actually benefit from \\ automatic process-level verification?}
\end{center}

In this paper, we address this question along four complementary dimensions: (i)~verification type (\emph{generative} versus \emph{scoring-based}), (ii)~verification granularity in MAS (verifying each \emph{agentic call} versus each \emph{full iteration}), (iii) context management strategies, and (iv) solvability. 
Our empirical study spans six representative MAS frameworks, two reasoning domains (mathematical problem solving and agentic search), and five distinct evaluators, enabling a systematic and controlled examination of when, how, and to what extent process-level verification benefits multi-agent systems.
Through this analysis, we identify both consistent patterns and high-variance behaviors across settings, clarifying the practical limits and trade-offs of applying process verification to MAS.

To facilitate reproducible research and establish a standardized testbed for the community, we introduce a universally applicable protocol, \textbf{MAS}-\textbf{Pro}cess \textbf{Ve}rification (\textbf{\ours}), for integrating process-level verification into MAS test-time scaling, as demonstrated in \Figref{fig:main}.
\ours is modular and extensible, operating as a plug-and-play wrapper for \emph{any} off-the-shelf MAS framework and \emph{any} verification method, and requiring only \emph{minimal} code modifications.

\textbf{Main Observations.}\quad
{We present the first systematic study of the strengths and limitations of process verification in MAS, spanning multiple dimensions including verification types, granularity, context management, and problem solvability. 

\begin{itemize}[nosep,leftmargin=18pt]
    \item \textbf{Verification type}: Process verification in MAS does not consistently yield performance improvements across verification types and often exhibits high variance, \blue{}{with judge-based verifiers dominating the reward-based verifiers.} This highlights the intrinsic difficulty of reliably evaluating partial multi-agent trajectories.
    
    \item \textbf{Verification Granularity}:
    While no single verification granularity universally dominates across MAS frameworks, individual MAS often exhibit a clear preference between agent-level and iteration-level verification.

    \item \textbf{Context Management Strategies}: Comparable performance between only step responses vs context + step(raw) responses, with summarization of responses leading the performance gains in long-context MAS. 
    
    \item \textbf{Solvability}: A question-wise analysis on mathematical reasoning benchmarks shows that the process verification with optimized settings can improve stability, but rarely recovers instances that are fundamentally unsolvable by MAS.
\end{itemize}

\begin{figure*}[t]
    \centering
    \includegraphics[width=1\linewidth]{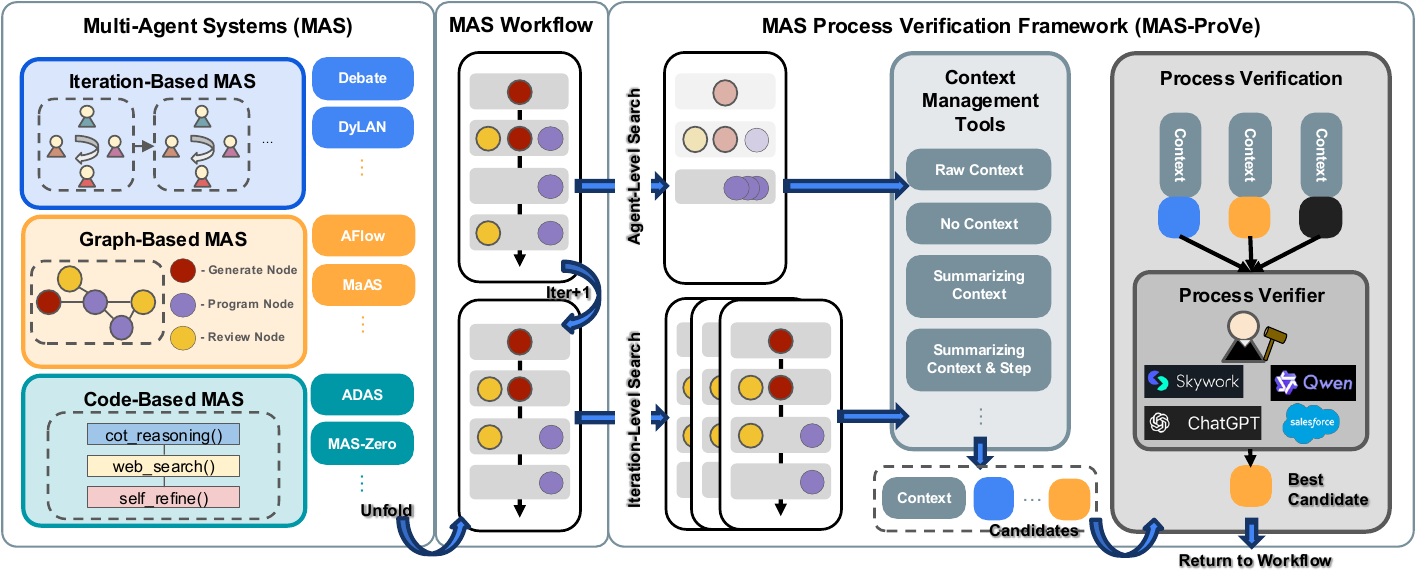}
    \vspace{-2em}
    \caption{
    \textbf{Overview of Multi-Agent System Process Verification Framework~(\ours).}
    The execution of \emph{any} multi-agent system (MAS) can be abstracted into a unified workflow, where stages are executed sequentially and atomic LLM calls within each stage may run in parallel.
    \ours systematically evaluates the integration of MAS with process verification by performing parallel search at two complementary granularities: \emph{agent-level} and \emph{iteration-level}, while remaining agnostic to the choice of process verifier.
    Prior to verification, customizable \emph{context-management strategies} can be applied to partial multi-agent trajectories, enabling flexible control over the information passed to the verifier and the selection of the best candidate to return to the workflow.
    }
    \label{fig:main}
\end{figure*}

\section{Background}
\label{sec:bg}
In this section, we establish the formal framework that governs our experimental design. Rather than focusing on specific models or datasets, we define the general abstractions for the two core components of our study: the taxonomy of Multi-Agent Systems (\Secref{sec:bg_mas}) and the classes of Automatic Verification (\Secref{sec:bg_verification}). Finally, in \Secref{sec:bg_marrying}, we unify these components to define the specific design space of our analysis—granularity, context, and search strategy, which serves as the theoretical basis for the concrete instantiations detailed in the experimental setup.

\subsection{MAS Overview}
\label{sec:bg_mas}
A Multi-Agent System (MAS) consists of multiple LLM-based agents, each equipped with its own context and (sub-)goals, coordinated as a system to perform a given task. We denote the resulting system-level behavior to an input query a \textit{reasoning trajectory}, where the output of each sub-agent can be considered a ``step'' towards the final output. While a general concept, a variety of recent works~\citep{liu2024dynamicllmpowered,chen2024agentverse,zhang2025aflow} have proposed different instantiations of MAS. Broadly, we consider two classes of MAS: (i) \textit{fixed architecture} systems, where the topology of agents and communication edges remains fixed for all problems, and (ii) \textit{adaptive architecture} systems, where the MAS topology may change based on a per-query basis. MAS belonging to the former may either be human designed MAS topologies (e.g., LLM-debate) or instantiated by iteratively searching for the best MAS structure on a small validation set (i.e., ``training''). In the latter, MAS tend to self-evolve over multiple iterations at test-time, typically using self-reflection to update agent roles and communication edges.

\subsection{Automatic Verification}
\label{sec:bg_verification}
To meet demands for scalable verification, automatic verifiers have been deployed to provide feedback signals about output and intermediate state correctness in reasoning settings~\citep{snell2024scaling,zhou2025evaluating,shao2025deepseekmath}. There are three primary classes of verifiers: (i) \textit{reward model~(RMs)}, which produce a continuous score for an input and terminal output state, (ii) \textit{process reward models~(PRMs)}, which produce continuous scores for intermediate states, measuring likelihood of reaching a correct outcome from a given state, and (iii) \textit{generative verifiers~(LLM-as-a-Judge)}, which outputs feedback and evaluation in natural language. For LLM-as-a-Judge baselines, we analyze both general purpose LLMs as judges and specialized, state-of-the-art finetuned verifier models.

\subsection{Marrying MAS and Verification}
\label{sec:bg_marrying}
Integrating verification into MAS settings gives rise to a search-based optimization problem that we seek to analyze, i.e., yielding multiple execution/trajectories in parallel and then letting the verifier to pick the most promising candidate. 
We choose three crucial facets for analysis:

\textbf{Verification Types.}\quad
To broadly cover the dominant classes of automatic verifiers used in prior work, \ours includes all three verifier types in our analysis. To the best of our knowledge, existing verifiers are not explicitly trained on multi-agent trajectories; as a result, their application to MAS constitutes a strictly out-of-distribution setting.
Among these approaches, LLM-as-a-Judge is typically instantiated using general-purpose LLMs with strong generalization capabilities. In contrast, reward models—particularly process reward models—are commonly trained on single-agent, verifiable reasoning trajectories (e.g., mathematical problem solving). Consequently, we hypothesize that LLM-as-a-Judge is more naturally suited to MAS test-time scaling than reward-based alternatives.

\textbf{Verification Granularity.}\quad
Verification-based interventions can occur at different levels of granularity. We study two concrete levels, as shown in~\Figref{fig:main} (middle): the \textit{Agent-Level}, where the outputs of each sub-agent are explicitly verified, and the \textit{Iteration-Level}, where the MAS itself is evaluated as a whole. The former enables catching fine-grained errors before they corrupt other sub-agent behavior, but introduces significant overhead. The latter takes a more global perspective potentially at the cost of finding step-level errors.

\textbf{Context Management.}\quad
The reliability of process verification critically depends on how the reasoning state is represented to the verifier. In MAS, trajectories may span many interaction turns; naively providing the full raw history can introduce bias due to context saturation, while conditioning only on the current step ignores the state dependencies required to assess reasoning continuity.
To empirically identify a ``sweet spot'' in contextual conditioning, we evaluate four context specifications with increasing information density:
\textit{(i)~current-step only},
\textit{(ii)~summarize(context) + current step},
\textit{(iii)~summarize(context + current step)}, and
\textit{(iv)~full context + current step}.

\textbf{Solvability.}\quad 
To disentangle the effect of process verification from the intrinsic reasoning capability of the underlying agents, we analyze MAS performance through the lens of \emph{problem solvability}. A verification signal can guide the system toward a correct solution only if such an outcome lies within the agent’s generation horizon. Consequently, it is essential to first characterize the difficulty of the benchmarks on which the MAS operates.
We partition each benchmark into three difficulty levels, \emph{Easy}, \emph{Medium}, and \emph{Hard} based on the empirical pass rate of the unguided CoT(Chain-of-Thought) baseline, and introduce task-specific metrics to quantify how process verification interacts with solvability. Specifically, we consider:
\emph{(i) Performance Gain},
\emph{(ii) Evaluator Stability}, and
\emph{(iii) Intra-MAS Resurrection}.

\section{Experimental Setup}
\label{sec:experiments}

\textbf{Benchmarks.}\quad 
Our analysis focuses on {four representative reasoning domains: mathematics~(\textbf{AIME24} and \textbf{AIME25},~\cite{aime}), agentic assistance~(\textbf{GAIA}~\cite{mialon2023gaia}), coding~(\textbf{HumanEval}~\cite{chen2021evaluating}), and scientific reasoning~(\textbf{GPQA}~\cite{rein2023gpqa}).} To support MAS baselines that require training or tuning, we partition each benchmark into a validation set (20\%) and a test set (80\%), and reuse the same splits across all methods for fair comparison.

\textbf{MAS Baselines and Implementation Details.}\quad
We conduct a comprehensive evaluation on six state-of-the-art MAS frameworks with both \underline{\emph{fixed architectures}}, including LLM-Debate~(\textbf{Debate})~\cite{du2024improving}, Agentic Workflow~(\textbf{AFlow})~\cite{zhang2025aflow}, and Automated Design of Agentic Systems~(\textbf{ADAS})~\cite{hu2025automated}, 
and \underline{\emph{adaptive architectures}}, including Dynamic LLM-Powered Agent Network~(\textbf{DyLAN})~\cite{liu2024dynamicllmpowered}, Multi-agent Architecture Search~(\textbf{MaAS})~\cite{zhang2025multiagent}, and MAS with Zero Supervision~(\textbf{MAS-Zero})~\cite{Ke2025MASZero}. 

To balance cost efficiency and reasoning capability, we adopt \textbf{GPT-5-Mini} as the primary backbone model across all baselines. {Only for HumanEval, we use the \textbf{GPT-4.1-nano} model; this prevents the performance ceiling that stronger models would hit on this simpler task, allowing us to accurately observe the delta introduced by process verification.} For the agentic benchmark GAIA, all MAS frameworks are further equipped with a DuckDuckGo search tool via OpenAI’s tool-calling interface. This enables dynamic external information retrieval and facilitates a systematic evaluation of process verification granularity in tool-assisted reasoning settings.
Detailed configurations of the MAS baselines can be found in \appref{app:baselines}.

\textbf{Verifiers.}\quad 
We consider three main verifiers, one for each category: \textbf{Skywork-Reward-V2-Llama-3.1-8B}~\citep{liu2025skywork} as an RM, \textbf{Qwen2.5-Math-PRM-7B}~\citep{zhang2025lessons} as a PRM, and \textbf{GPT-5-Mini} as an LLM-as-Judge. In extended analysis, we also consider the weaker GPT-4o-Mini and the finetuned verifier \textbf{FARE-20B}~\citep{xu2025foundational}, which is trained from \textbf{GPT-OSS-20B}~\citep{agarwal2025gpt}.

\newcommand{\hlgreen}[1]{\colorbox{lightgreen}{#1}}
\newcommand{\hlred}[1]{\colorbox{lightred}{#1}}
\begin{table*}[t]
\caption{
    \textbf{Main Study of \ours on Verification Types and Granularity.}
    We investigate two core dimensions of process verification for multi-agent systems:
    \emph{(i)} the choice of process verifier (LLM-as-a-Judge, Reward Model, or Process Reward Model), and
    \emph{(ii)} the granularity at which verification is applied.
    All experiments are repeated three times, and we report the mean and standard deviation. 
    Values reported under the \emph{No ProVe} row for each MAS correspond to the \emph{baseline performance.} 
    Higher values indicate better performance throughout.
    \textbf{Boldface} denotes the best result in each setting. \hlgreen{Green cells} indicate improvements over the baseline, while \hlred{red cells} indicate degradations relative to the baseline.
}
\vspace{-1em}
\begin{center}
\resizebox{1\linewidth}{!}{%
\setlength{\tabcolsep}{6pt}
\begin{tabular}{lccccc ccc ccc ccc }
	\toprule[0.12em]
	\multirow{2}{*}[-0.25em]{\textbf{Method}} 
    & \multirow{2}{*}[-0.25em]{\textbf{MAS-Arch}} 
    & \multirow{2}{*}[-0.25em]{\textbf{ProVe Level}}
    & \multicolumn{3}{c}{{\textbf{AIME24}}} 
    & \multicolumn{3}{c}{{\textbf{AIME25}}} 
    & \multicolumn{3}{c}{{\textbf{GAIA}}} 
    \\
     
    \cmidrule(lr){4-6} \cmidrule(lr){7-9} \cmidrule(lr){10-12}
    & & & Judge
    & RM
    & PRM
    & Judge
    & RM
    & PRM
    & Judge
    & RM
    & PRM
    \\
    \midrule

    \multirow{3}{*}[-0.25em]{Debate}
    & \multirow{3}{*}[-0.25em]{Fixed} 
    & No ProVe
    & 
    & 74.13\scriptsize{$\pm$3.44}
    & 
    & 
    & 61.87\scriptsize{$\pm$3.21}
    & 
    & 
    & 31.07\scriptsize{$\pm$2.07}
    \\

    & 
    & Agent
    & \cellcolor{lightgreen}{\textbf{81.11\scriptsize{$\pm$1.92}}}
    & \cellcolor{lightred}{71.67\scriptsize{$\pm$1.44}}
    & \cellcolor{lightred}{68.10\scriptsize{$\pm$4.50}}
    & \cellcolor{lightgreen}{\textbf{62.97\scriptsize{$\pm$3.38}}}
    & \cellcolor{lightred}{54.45\scriptsize{$\pm$3.85}}
    & \cellcolor{lightred}{57.12\scriptsize{$\pm$8.85}}
    & \cellcolor{lightred}{\textbf{30.28\scriptsize{$\pm$1.51}}}
    & \cellcolor{lightred}{25.05\scriptsize{$\pm$2.33}}
    & \cellcolor{lightred}{26.20\scriptsize{$\pm$0.14}}
    \\

    & 
    & Iteration
    & \cellcolor{lightred}{72.67\scriptsize{$\pm$5.97}}
    & \cellcolor{lightgreen}{\textbf{81.11\scriptsize{$\pm$3.85}}}
    & \cellcolor{lightred}{70.34\scriptsize{$\pm$3.07}}
    & \cellcolor{lightred}{53.33\scriptsize{$\pm$8.82}}
    & \cellcolor{lightred}{\textbf{57.78\scriptsize{$\pm$7.70}}}
    & \cellcolor{lightred}{54.59\scriptsize{$\pm$2.18}}
    & \cellcolor{lightgreen}{\textbf{32.68\scriptsize{$\pm$3.11}}}
    & \cellcolor{lightred}{20.60\scriptsize{$\pm$1.21}}
    & \cellcolor{lightred}{23.20\scriptsize{$\pm$4.50}}
    \\

    \midrule

    \multirow{3}{*}[-0.25em]{AFlow}
    & \multirow{3}{*}[-0.25em]{Fixed} 
    & No ProVe
    & 
    & 65.67\scriptsize{$\pm$5.09}
    & 
    & 
    & 59.72\scriptsize{$\pm$6.37}
    &
    &
    & 11.50\scriptsize{$\pm$3.25}
    \\

    & 
    & Agent
    & \cellcolor{lightgreen}{\textbf{76.39\scriptsize{$\pm$6.36}}}
    & \cellcolor{lightgreen}{76.22\scriptsize{$\pm$6.59}}
    & \cellcolor{lightgreen}{74.17\scriptsize{$\pm$3.01}}
    & \cellcolor{lightred}{\textbf{58.33\scriptsize{$\pm$4.17}}}
    & \cellcolor{lightred}{58.32\scriptsize{$\pm$11.04}}
    & \cellcolor{lightred}{56.94\scriptsize{$\pm$8.68}}
    & \cellcolor{lightgreen}{\textbf{21.68\scriptsize{$\pm$2.08}}}
    & \cellcolor{lightgreen}{18.08\scriptsize{$\pm$5.52}}
    & \cellcolor{lightgreen}{14.62\scriptsize{$\pm$2.03}}
    \\

    & 
    & Iteration
    & \cellcolor{lightgreen}{\textbf{69.44\scriptsize{$\pm$8.67}}}
    & \cellcolor{lightred}{52.73\scriptsize{$\pm$10.52}}
    & \cellcolor{lightgreen}{61.11\scriptsize{$\pm$2.41}}
    & \cellcolor{lightgreen}{\textbf{60.39\scriptsize{$\pm$6.26}}}
    & \cellcolor{lightred}{54.16\scriptsize{$\pm$5.89}}
    & \cellcolor{lightred}{49.99\scriptsize{$\pm$4.18}}
    & \cellcolor{lightgreen}{18.62\scriptsize{$\pm$0.78}}
    & \cellcolor{lightgreen}{\textbf{21.64\scriptsize{$\pm$1.64}}}
    & \cellcolor{lightgreen}{13.85\scriptsize{$\pm$2.55}}
    \\

    \midrule

    \multirow{3}{*}[-0.25em]{ADAS}
    & \multirow{3}{*}[-0.25em]{Fixed} 
    & No ProVe
    &
    & 62.49\scriptsize{$\pm$4.19}
    &
    &
    & 54.17\scriptsize{$\pm$4.15}
    &
    &
    & 19.30\scriptsize{$\pm$0.92}
    \\

    & 
    & Agent
    & \cellcolor{lightgreen}{73.33\scriptsize{$\pm$12.83}}
    & \cellcolor{lightgreen}{\textbf{73.93\scriptsize{$\pm$8.26}}}
    & \cellcolor{lightgreen}{68.07\scriptsize{$\pm$9.64}}
    & \cellcolor{lightgreen}{58.33\scriptsize{$\pm$0.00}}
    & \cellcolor{lightgreen}{\textbf{61.10\scriptsize{$\pm$2.42}}}
    & \cellcolor{lightred}{48.90\scriptsize{$\pm$2.42}}
    & \cellcolor{lightgreen}{20.90\scriptsize{$\pm$2.50}}
    & \cellcolor{lightgreen}{\textbf{22.10\scriptsize{$\pm$2.50}}}
    & \cellcolor{lightgreen}{19.70\scriptsize{$\pm$4.33}}
    \\

    & 
    & Iteration
    & \cellcolor{lightgreen}{\textbf{70.83\scriptsize{$\pm$15.03}}}
    & \cellcolor{lightgreen}{66.67\scriptsize{$\pm$11.06}}
    & \cellcolor{lightgreen}{63.87\scriptsize{$\pm$6.36}}
    & \cellcolor{lightgreen}{\textbf{61.10\scriptsize{$\pm$4.85}}}
    & \cellcolor{lightgreen}{59.73\scriptsize{$\pm$4.79}}
    & \cellcolor{lightgreen}{54.23\scriptsize{$\pm$0.00}}
    & \cellcolor{lightgreen}{\textbf{22.90\scriptsize{$\pm$2.08}}}
    & \cellcolor{lightgreen}{19.70\scriptsize{$\pm$1.83}}
    & \cellcolor{lightgreen}{18.53\scriptsize{$\pm$3.01}}
    \\

    \midrule

    \multirow{3}{*}[-0.25em]{MaAS}
    & \multirow{3}{*}[-0.25em]{Adaptive} 
    &  No ProVe
    &  
    & 76.39\scriptsize{$\pm$6.36}
    & 
    &  
    & 58.33\scriptsize{$\pm$7.22}
    & 
    & 
    & 20.97\scriptsize{$\pm$4.90}
    \\

    & 
    & Agent
    & \cellcolor{lightgreen}{80.55\scriptsize{$\pm$8.67}}
    & \cellcolor{lightgreen}{\textbf{84.03\scriptsize{$\pm$6.73}}}
    & \cellcolor{lightgreen}{83.33\scriptsize{$\pm$4.17}}
    & \cellcolor{lightgreen}{\textbf{61.09\scriptsize{$\pm$4.83}}}
    & \cellcolor{lightred}{52.08\scriptsize{$\pm$2.94}}
    & \cellcolor{lightgreen}{59.70\scriptsize{$\pm$2.42}}
    & \cellcolor{lightgreen}{\textbf{26.89\scriptsize{$\pm$5.11}}}
    & \cellcolor{lightgreen}{25.36\scriptsize{$\pm$6.41}}
    & \cellcolor{lightgreen}{24.70\scriptsize{$\pm$0.86}}
    \\

    & 
    & Iteration
    & \cellcolor{lightgreen}{77.22\scriptsize{$\pm$10.48}}
    & \cellcolor{lightgreen}{\textbf{86.11\scriptsize{$\pm$6.37}}}
    & \cellcolor{lightgreen}{81.94\scriptsize{$\pm$6.36}}
    & \cellcolor{lightgreen}{\textbf{68.06\scriptsize{$\pm$6.36}}}
    & \cellcolor{lightgreen}{63.89\scriptsize{$\pm$2.41}}
    & \cellcolor{lightgreen}{63.88\scriptsize{$\pm$8.68}}
    & \cellcolor{lightgreen}{\textbf{31.21\scriptsize{$\pm$3.60}}}
    & \cellcolor{lightgreen}{27.29\scriptsize{$\pm$2.55}}
    & \cellcolor{lightgreen}{24.26\scriptsize{$\pm$0.97}}
    \\
    
    \midrule

    \multirow{3}{*}[-0.25em]{DyLAN}
    & \multirow{3}{*}[-0.25em]{Adaptive} 
    &No ProVe
    &
    & 81.11\scriptsize{$\pm$5.09}
    &
    &
    & 65.55\scriptsize{$\pm$5.09}
    &
    &
    & 19.37\scriptsize{$\pm$0.95}
    \\

    & 
    & Agent
    & \cellcolor{lightgreen}{\textbf{85.56\scriptsize{$\pm$1.93}}}
    & \cellcolor{lightred}{76.67\scriptsize{$\pm$0.00}}
    & \cellcolor{lightred}{77.78\scriptsize{$\pm$3.85}}
    & \cellcolor{lightgreen}{70.00\scriptsize{$\pm$3.85}}
    & \cellcolor{lightgreen}{\textbf{72.22\scriptsize{$\pm$5.09}}}
    & \cellcolor{lightgreen}{67.78\scriptsize{$\pm$3.85}}
    & \cellcolor{lightred}{18.12\scriptsize{$\pm$2.02}}
    & \cellcolor{lightgreen}{\textbf{19.41\scriptsize{$\pm$1.68}}}
    & \cellcolor{lightred}{16.50\scriptsize{$\pm$1.37}}
    \\

    & 
    & Iteration
    & \cellcolor{lightgreen}{\textbf{84.44\scriptsize{$\pm$1.93}}}
    & \cellcolor{lightred}{78.89\scriptsize{$\pm$1.92}}
    & \cellcolor{lightgreen}{82.22\scriptsize{$\pm$10.71}} 
    & \cellcolor{lightgreen}{71.11\scriptsize{$\pm$3.85}}
    & \cellcolor{lightgreen}{\textbf{73.33\scriptsize{$\pm$5.78}}}
    & \cellcolor{lightgreen}{67.77\scriptsize{$\pm$8.39}}
    & \cellcolor{lightred}{\textbf{15.50\scriptsize{$\pm$1.41}}}
    & \cellcolor{lightred}{14.83\scriptsize{$\pm$1.53}}
    & \cellcolor{lightred}{9.70\scriptsize{$\pm$1.37}}
    \\

    \midrule

    \multirow{3}{*}[-0.25em]{MAS-Zero}
    & \multirow{3}{*}[-0.25em]{Adaptive} 
    & No ProVe
    &
    & 37.78\scriptsize{$\pm$1.92}
    &
    &
    & 35.56\scriptsize{$\pm$1.93}
    &
    & 
    & 10.47\scriptsize{$\pm$0.68}
    \\

    & 
    & Agent
    & \cellcolor{lightgreen}{\textbf{47.78\scriptsize{$\pm$1.92}}}
    & \cellcolor{lightgreen}{38.89\scriptsize{$\pm$1.92}}
    & \cellcolor{lightred}{33.34\scriptsize{$\pm$5.77}}
    & \cellcolor{lightgreen}{\textbf{42.22\scriptsize{$\pm$3.85}}}
    & \cellcolor{lightred}{28.87\scriptsize{$\pm$5.10}}
    & \cellcolor{lightred}{20.00\scriptsize{$\pm$3.85}}
    & \cellcolor{lightgreen}{17.36\scriptsize{$\pm$0.97}}
    & \cellcolor{lightgreen}{\textbf{19.14\scriptsize{$\pm$1.06}}}
    & \cellcolor{lightred}{10.14\scriptsize{$\pm$0.62}}
    \\

    & 
    & Iteration
    & \cellcolor{lightgreen}{\textbf{50.00\scriptsize{$\pm$3.33}}}
    & \cellcolor{lightgreen}{45.00\scriptsize{$\pm$2.36}}
    & \cellcolor{lightred}{37.78\scriptsize{$\pm$1.92}}
    & \cellcolor{lightred}{\textbf{27.78\scriptsize{$\pm$1.92}}}
    & \cellcolor{lightred}{25.56\scriptsize{$\pm$5.09}}
    & \cellcolor{lightred}{25.56\scriptsize{$\pm$8.39}}
    & \cellcolor{lightgreen}{\textbf{17.95\scriptsize{$\pm$1.11}}}
    & \cellcolor{lightgreen}{14.94\scriptsize{$\pm$0.62}}
    & \cellcolor{lightred}{8.20\scriptsize{$\pm$0.71}}
    \\

    \bottomrule[0.12em]
    \end{tabular}
 }
\end{center}
\label{tab:main}
\vspace{-0.5em}
\end{table*}

\begin{table*}[t]
\caption{
    \textbf{Main Study of \ours on Verification Types and Granularity (on Coding and Scientific Reasoning).}
    Higher values indicate better performance throughout.
    \textbf{Boldface} denotes the best result in each setting. \hlgreen{Green cells} indicate improvements over the baseline, while \hlred{red cells} indicate degradations relative to the baseline.
}
\vspace{-1em}
\begin{center}
\resizebox{1\linewidth}{!}{%
\setlength{\tabcolsep}{15pt}
\begin{tabular}{lccccc ccc ccc }
	\toprule[0.12em]
	\multirow{2}{*}[-0.25em]{\textbf{Method}} 
    & \multirow{2}{*}[-0.25em]{\textbf{MAS-Arch}} 
    & \multirow{2}{*}[-0.25em]{\textbf{ProVe Level}}
    & \multicolumn{3}{c}{{\textbf{HumanEval}}} 
    & \multicolumn{3}{c}{{\textbf{GPQA}}} 
    \\
     
    \cmidrule(lr){4-6} \cmidrule(lr){7-9} 
    & & & Judge
    & RM
    & PRM
    & Judge
    & RM
    & PRM
    \\
    \midrule

    \multirow{3}{*}[-0.25em]{Debate}
    & \multirow{3}{*}[-0.25em]{Fixed} 
    & No ProVe
    &
    & 76.40\scriptsize{$\pm$1.56}
    &
    &
    & 59.13\scriptsize{$\pm$3.50}
    & 
    \\

    & 
    & Agent
    &  \cellcolor{lightgreen}\textbf{78.35\scriptsize{$\pm$0.49}}
    & \cellcolor{lightred}76.10\scriptsize{$\pm$0.17}
    & \cellcolor{lightgreen}78.04\scriptsize{$\pm$0.00}
    & \cellcolor{lightgreen}\textbf{{66.82\scriptsize{$\pm$3.36}}}
    & \cellcolor{lightred}{52.48\scriptsize{$\pm$2.79}}
    & \cellcolor{lightred}{55.25\scriptsize{$\pm$0.28}}
    \\

    & 
    & Iteration
    &  \cellcolor{lightgreen}\textbf{77.72\scriptsize{$\pm$0.45}}
    & \cellcolor{lightred}76.20\scriptsize{$\pm$0.85}
    & \cellcolor{lightgreen}77.02\scriptsize{$\pm$0.71}
    & \cellcolor{lightgreen}\textbf{{70.95\scriptsize{$\pm$0.35}}}
    & \cellcolor{lightgreen}{69.19\scriptsize{$\pm$0.71}}
    & \cellcolor{lightred}{58.53\scriptsize{$\pm$5.78}}
    \\

    \midrule

    \multirow{3}{*}[-0.25em]{AFlow}
    & \multirow{3}{*}[-0.25em]{Fixed} 
    & No ProVe
    &
    & 88.78\scriptsize{$\pm$1.17}
    &
    &
    & 46.33\scriptsize{$\pm$0.96}
    
    \\

    & 
    & Agent
    & \cellcolor{lightgreen}{\textbf{89.80\scriptsize{$\pm$1.19}}}
    & \cellcolor{lightred}{88.14\scriptsize{$\pm$0.51}}
    & \cellcolor{lightred}{73.66\scriptsize{$\pm$4.86}}
    & \cellcolor{lightgreen}{\textbf{54.93\scriptsize{$\pm$2.65}}}
    & \cellcolor{lightred}{44.24\scriptsize{$\pm$1.32}}
    & \cellcolor{lightred}{39.72\scriptsize{$\pm$1.78}}
    \\

    & 
    & Iteration
    & \cellcolor{lightgreen}{92.58\scriptsize{$\pm$0.48}}
    & \cellcolor{lightgreen}{\textbf{92.75\scriptsize{$\pm$0.54}}}
    & \cellcolor{lightgreen}{89.28\scriptsize{$\pm$1.11}}
    & \cellcolor{lightgreen}{\textbf{52.20\scriptsize{$\pm$3.55}}}
    & \cellcolor{lightred}{40.56\scriptsize{$\pm$2.22}}
    & \cellcolor{lightred}{37.73\scriptsize{$\pm$1.78}}
    \\

    \midrule

    \multirow{3}{*}[-0.25em]{ADAS}
    & \multirow{3}{*}[-0.25em]{Fixed} 
    & No ProVe
    & 
    & 95.42\scriptsize{$\pm$3.12}
    &
    &
    & 66.54\scriptsize{$\pm$1.83}
    \\

    & 
    & Agent
    & \cellcolor{lightred}{94.82\scriptsize{$\pm$2.41}}
    & \cellcolor{lightred}{\textbf{95.03\scriptsize{$\pm$3.83}}}
    & \cellcolor{lightred}{94.08\scriptsize{$\pm$4.37}}
    & \cellcolor{lightgreen}{68.94\scriptsize{$\pm$3.64}}
    & \cellcolor{lightgreen}\textbf{{69.16\scriptsize{$\pm$4.31}}}
    & \cellcolor{lightred}{60.79\scriptsize{$\pm$4.82}}
    \\

    & 
    & Iteration
    & \cellcolor{lightgreen}{\textbf{96.38\scriptsize{$\pm$2.63}}}
    & \cellcolor{lightred}{95.20\scriptsize{$\pm$2.88}}
    & \cellcolor{lightgreen}{96.20\scriptsize{$\pm$1.65}}
    & \cellcolor{lightgreen}{\textbf{69.54\scriptsize{$\pm$2.90}}}
    & \cellcolor{lightred}{55.23\scriptsize{$\pm$3.76}}
    & \cellcolor{lightred}{51.65\scriptsize{$\pm$2.31}}
    \\

    \midrule

    \multirow{3}{*}[-0.25em]{MaAS}
    & \multirow{3}{*}[-0.25em]{Adaptive} 
    &  No ProVe
    &
    & 96.18\scriptsize{$\pm$1.08}
    &
    &
    & 66.23\scriptsize{$\pm$1.44}
    \\

    & 
    & Agent
    & \cellcolor{lightred}{{95.93\scriptsize{$\pm$0.44}}}
    & \cellcolor{lightred}{{95.80\scriptsize{$\pm$2.70}}}
    & \cellcolor{lightgreen}{{\textbf{96.18\scriptsize{$\pm$0.00}}}}
    & \cellcolor{lightgreen}{\textbf{{67.90\scriptsize{$\pm$0.86}}}}
    & \cellcolor{lightgreen}{{66.66\scriptsize{$\pm$0.89}}}
    & \cellcolor{lightred}{{65.09\scriptsize{$\pm$1.33}}}
    \\

    & 
    & Iteration
    & \cellcolor{lightgreen}{\textbf{97.33\scriptsize{$\pm$0.54}}}
    & \cellcolor{lightgreen}{96.56\scriptsize{$\pm$0.54}}
    & \cellcolor{lightgreen}{96.18\scriptsize{$\pm$2.16}}
    & \cellcolor{lightgreen}{\textbf{71.06\scriptsize{$\pm$2.67}}}   
    & \cellcolor{lightgreen}{68.33\scriptsize{$\pm$1.46}}
    & \cellcolor{lightred}{61.62\scriptsize{$\pm$0.87}}
    \\
    
    \midrule

    \multirow{3}{*}[-0.25em]{DyLAN}
    & \multirow{3}{*}[-0.25em]{Adaptive} 
    &No ProVe
    & 
    & 90.23\scriptsize{$\pm$0.00}
    &
    &
    & 68.18\scriptsize{$\pm$1.43}
    \\

    & 
    & Agent
    & \cellcolor{lightgreen}{\textbf{93.65\scriptsize{$\pm$0.35}}}
    & \cellcolor{lightgreen}91.16\scriptsize{$\pm$1.29}
    & \cellcolor{lightred}89.72\scriptsize{$\pm$0.74}
    & \cellcolor{lightgreen}{\textbf{78.76\scriptsize{$\pm$2.91}}}
    & \cellcolor{lightgreen}76.46\scriptsize{$\pm$0.29}
    & \cellcolor{lightred}67.92\scriptsize{$\pm$2.50}
    \\

    & 
    & Iteration
    & \cellcolor{lightgreen}{\textbf{94.20\scriptsize{$\pm$0.42}}}
    & \cellcolor{lightgreen}92.38\scriptsize{$\pm$0.43}
    & \cellcolor{lightgreen}93.62\scriptsize{$\pm$0.40}
    & \cellcolor{lightgreen}{\textbf{72.20\scriptsize{$\pm$1.41}}}
    & \cellcolor{lightgreen}69.44\scriptsize{$\pm$1.77}
    & \cellcolor{lightgreen}68.68\scriptsize{$\pm$1.43}
    \\

    \midrule

    \multirow{3}{*}[-0.25em]{MAS-Zero}
    & \multirow{3}{*}[-0.25em]{Adaptive} 
    & No ProVe
    & 
    & 53.63\scriptsize{$\pm$3.00}
    &
    &
    & 60.72\scriptsize{$\pm$3.50}
    \\

    & 
    & Agent
    & \cellcolor{lightgreen}\textbf{55.75\scriptsize{$\pm$0.00}}
    & \cellcolor{lightgreen}53.63\scriptsize{$\pm$1.62}
    & \cellcolor{lightred}50.91\scriptsize{$\pm$2.57}
    & \cellcolor{lightgreen}{\textbf{68.00\scriptsize{$\pm$1.24}}}
    & \cellcolor{lightred}61.41\scriptsize{$\pm$1.13}
    & \cellcolor{lightred}55.99\scriptsize{$\pm$0.41}
    \\

    & 
    & Iteration
    & \cellcolor{lightgreen}\textbf{54.23\scriptsize{$\pm$2.14}}
    & \cellcolor{lightred}51.81\scriptsize{$\pm$1.29}
    & \cellcolor{lightred}52.72\scriptsize{$\pm$0.00}
    & \cellcolor{lightgreen}{61.31\scriptsize{$\pm$2.86}}
    &\cellcolor{lightgreen}\textbf{64.06\scriptsize{$\pm$0.36}}
    &\cellcolor{lightred} 53.77\scriptsize{$\pm$1.48}
    \\

    \bottomrule[0.12em]
    \end{tabular}
 }
\end{center}
\label{tab:main-coding-science}
\vspace{-0.5em}
\end{table*}

\textbf{Integrating Verifiers into MAS via Search.}\quad
Given verification scores produced by the chosen evaluators, the system must determine how to advance the reasoning process. We adopt a \emph{greedy best-first search} strategy at the specified verification granularity for every MAS framework to ensure a scalable, effective, and fair comparison. At each state, the MAS generates $N{=}3$ candidate continuations and selects the highest-scoring/ranking candidate according to the verifier to define the next state. This procedure effectively transforms the MAS from a purely stochastic generator into a verifier-guided search process. 
{To further enrich the search heuristics and provide a more comprehensive exploration of this bounded space, we additionally implement Beam Search (beam width${=}$3) on top of the Debate framework using our optimal Generative Judge, as detailed in \appref{app:beam_search}.}

\textbf{\ours Framework.}\quad 
A broad overview of our framework is illustrated in \Figref{fig:main}. 
\emph{Conceptually,} the execution of \emph{any} multi-agent system can be abstracted into a unified workflow, where stages are executed sequentially and atomic LLM calls within each stage may run in parallel.
\ours systematically evaluates the integration of MAS with process verification by performing parallel search at two complementary granularities: \emph{agent-level} and \emph{iteration-level}, while remaining agnostic to the choice of process verifier.
Prior to verification, customizable \emph{context-management strategies} can be applied to partial multi-agent trajectories, enabling flexible control over the information passed to the verifier and the selection of the best candidate to return to the workflow.

\emph{Practically,} the \ours framework is implemented as an isolated \textit{client-middleware-server} architecture:
\begin{itemize}[nosep,leftmargin=18pt]
    \item \textbf{Client side.}
    The client executes the original MAS agentic logic \textit{without modification,} as at each reasoning step, the wrapper function automatically executes the original LLM call for $N$ times and interfaces with the middleware to obtain the selected continuation and resumes execution along the chosen path.
    
    \item \textbf{Middleware side.}  
    A context-management tool intercepts the agent’s execution, retrieves the accumulated context from the MAS, and re-organizes the context. Then these partial trajectories are forwarded to the server for evaluation.
    
    \item \textbf{Server side.}  
    The server hosts a pool of verifiers that score and rank the received partial trajectories. The resulting rankings are returned to the client via the middleware, which selects the highest-ranked continuation and enables the MAS to proceed.
\end{itemize}
{\textbf{Remark on Computational Overhead.}\quad 
\ours's systematic study inherently incurs additional computational overhead and is not intended to reflect the practical integration of process verification into MAS. A detailed breakdown of the wall-clock time is provided in \appref{app:computational_overhead}.}

\section{Main Results}
\label{sec:experiments-main}

\begin{table*}[t]
\caption{
    \textbf{Study of \ours on LLM-as-a-Judge as Verifiers.}
    \textbf{Boldface} denotes the best performance. 
}
\vspace{-1em}
\begin{center}
\resizebox{1\linewidth}{!}{%
\setlength{\tabcolsep}{6.5pt}
\begin{tabular}{lccccc ccc ccc ccc }
	\toprule[0.12em]
	\multirow{2}{*}[-0.25em]{\textbf{Method}} 
    & \multicolumn{3}{c}{{\textbf{AIME24}}} 
    
    & \multicolumn{3}{c}{{\textbf{AIME25}}} 
    & \multicolumn{3}{c}{{\textbf{GAIA}}} 
    \\
     
    \cmidrule(lr){2-4} \cmidrule(lr){5-7} \cmidrule(lr){8-10}
    & GPT-4o-mini
    & GPT-5-mini
    & FARE-20B
    & GPT-4o-mini
    & GPT-5-mini
    & FARE-20B
    & GPT-4o-mini
    & GPT-5-mini
    & FARE-20B
    \\
    \midrule

    {Debate}
    & 72.23\scriptsize{$\pm$3.87}
    & \textbf{81.11\scriptsize{$\pm$1.92}}
    & 76.67\scriptsize{$\pm$1.65}
    & 54.47\scriptsize{$\pm$6.32}
    & 62.97\scriptsize{$\pm$3.38}
    & \textbf{64.43\scriptsize{$\pm$5.38}}
    & 26.85\scriptsize{$\pm$4.42}
    & \textbf{30.28\scriptsize{$\pm$1.51}}
    & 27.87\scriptsize{$\pm$2.66}
    \\

    {AFlow
}
    & 76.39\scriptsize{$\pm$8.67}
    & 76.39\scriptsize{$\pm$6.36}
    & \textbf{81.94\scriptsize{$\pm$4.82}}
    & 58.33\scriptsize{$\pm$4.17}
    & 58.33\scriptsize{$\pm$4.17}
    & \textbf{61.10\scriptsize{$\pm$2.42}}
    & 19.27\scriptsize{$\pm$4.17}
    & \textbf{21.68\scriptsize{$\pm$2.08}}
    & 16.55\scriptsize{$\pm$0.85}
    \\

    {ADAS}
    & 69.45\scriptsize{$\pm$4.81}
    & \textbf{73.33\scriptsize{$\pm$12.83}}
    & 68.07\scriptsize{$\pm$9.64}
    & 63.87\scriptsize{$\pm$6.36}
    & 58.33\scriptsize{$\pm$0.00}
    & \textbf{66.66\scriptsize{$\pm$4.15}}
    & 20.50\scriptsize{$\pm$2.08}
    & 20.90\scriptsize{$\pm$2.50}
    & \textbf{22.50\scriptsize{$\pm$4.85}}
    \\
    
    {MAAS}
    & 83.31\scriptsize{$\pm$8.30}
    & 80.55\scriptsize{$\pm$8.67}
    & \textbf{84.73\scriptsize{$\pm$6.36}}
    & \textbf{62.50\scriptsize{$\pm$4.17}}
    & 61.09\scriptsize{$\pm$4.83}
    & 58.33\scriptsize{$\pm$4.17}
    & 26.50\scriptsize{$\pm$3.19}
    & 26.89\scriptsize{$\pm$5.11}
    & \textbf{28.31\scriptsize{$\pm$0.85}}
    
    \\

    {DyLAN}
    & 83.33\scriptsize{$\pm$3.34}
    & 85.56\scriptsize{$\pm$1.93}
    & \textbf{86.67\scriptsize{$\pm$3.34}}
    & \textbf{73.33\scriptsize{$\pm$5.78}}
    & 70.00\scriptsize{$\pm$3.85}
    & \textbf{73.33\scriptsize{$\pm$3.34}}
    & \textbf{19.37\scriptsize{$\pm$0.95}}
    & 18.12\scriptsize{$\pm$2.02}
    & 18.37\scriptsize{$\pm$0.98}
    \\

    {MAS-Zero}
    & 42.22\scriptsize{$\pm$5.09}
    & 47.78\scriptsize{$\pm$1.92}
    & \textbf{51.11\scriptsize{$\pm$1.92}}
    & 30.00\scriptsize{$\pm$3.33}
    & \textbf{42.22\scriptsize{$\pm$3.85}}
    & 38.89\scriptsize{$\pm$1.92}
    & 16.62\scriptsize{$\pm$1.48}
    & 17.36\scriptsize{$\pm$0.97}
    & \textbf{21.15\scriptsize{$\pm$2.72}}
    \\ 
     
    \bottomrule[0.12em]
    \end{tabular}
 }
\end{center}
\label{tab:main-judge}
\vspace{-1em}
\end{table*}

\begin{table*}[t]
\caption{
    \textbf{Comparing Different Context Management Strategies for Process Evaluation.} 
    \textbf{Boldface} denotes the best performance.
}
\vspace{-1em}
\begin{center}
\resizebox{1\linewidth}{!}{%
\setlength{\tabcolsep}{3pt}
\begin{tabular}{lccccc ccc ccc ccc }
	\toprule[0.12em]
	\multirow{2}{*}[-0.25em]{\textbf{Method}} 
    & \multicolumn{4}{c}{{\textbf{AIME24}}} 
    & \multicolumn{4}{c}{{\textbf{AIME25}}} 
    & \multicolumn{3}{c}{{\textbf{GAIA}}} 
    \\
     
    \cmidrule(lr){2-5} \cmidrule(lr){6-9} \cmidrule(lr){10-12}
    & Step-Only
    & Raw
    & Summary
    & Cont. Summary
    & Step-Only
    & Raw
    & Summary
    & Cont. Summary
    & Step-Only
    & Raw
    & Summary
    \\
    \midrule

    {Debate}
    & \textbf{81.11\scriptsize{$\pm$1.92}}
    & \textbf{81.11\scriptsize{$\pm$1.92}}
    & 77.78\scriptsize{$\pm$3.85}
    & -
    & 62.97\scriptsize{$\pm$3.38}
    & 62.97\scriptsize{$\pm$3.38}
    & \textbf{70.00\scriptsize{$\pm$6.67}}
    & -
    & \textbf{30.28\scriptsize{$\pm$1.51}}
    & \textbf{30.28\scriptsize{$\pm$1.51}}
    & 24.78\scriptsize{$\pm$0.47}
    \\

    {AFlow} 
    & 79.16\scriptsize{$\pm$4.17}
    & 76.39\scriptsize{$\pm$6.36}
    & 76.39\scriptsize{$\pm$6.36}
    & \textbf{80.55\scriptsize{$\pm$4.81}}
    & 65.28\scriptsize{$\pm$6.37}
    & 58.33\scriptsize{$\pm$4.17}
    & \textbf{69.44\scriptsize{$\pm$2.40}}
    & 66.67\scriptsize{$\pm$4.17}
    & 18.05\scriptsize{$\pm$2.11}
    & \textbf{21.68\scriptsize{$\pm$2.08}}
    & 16.86\scriptsize{$\pm$3.18}
    \\

    {ADAS}
    & 72.23\scriptsize{$\pm$6.37}
    & 73.33\scriptsize{$\pm$12.83}
    & 70.83\scriptsize{$\pm$11.00}
    & \textbf{73.63\scriptsize{$\pm$12.01}}
    & \textbf{66.67\scriptsize{$\pm$0.00}}
    & 58.33\scriptsize{$\pm$0.00}
    & 65.27\scriptsize{$\pm$4.79}
    & \textbf{66.67\scriptsize{$\pm$0.00}}
    & \textbf{22.50\scriptsize{$\pm$1.83}}
    & 20.90\scriptsize{$\pm$2.50}
    & 22.10\scriptsize{$\pm$3.02}
    \\
    
    {MAAS}
    & 81.94\scriptsize{$\pm$2.41}
    & 80.55\scriptsize{$\pm$8.67}
    & \textbf{84.72\scriptsize{$\pm$4.82}}
    & \textbf{84.72\scriptsize{$\pm$6.37}}
    & 65.28\scriptsize{$\pm$13.39}
    & 61.09\scriptsize{$\pm$4.83}
    & 62.49\scriptsize{$\pm$4.19}
    & \textbf{68.06\scriptsize{$\pm$2.40}}
    & 24.10\scriptsize{$\pm$1.70}
    & \textbf{26.89\scriptsize{$\pm$5.11}}
    & 23.45\scriptsize{$\pm$2.62}
    \\

    {DyLAN}
    &\textbf{ 86.67\scriptsize{$\pm$3.34}}
    & 85.56\scriptsize{$\pm$1.93}
    & 82.22\scriptsize{$\pm$5.09}
    & \textbf{ 86.67\scriptsize{$\pm$3.34}}
    & \textbf{80.00\scriptsize{$\pm$6.67}}
    & 70.00\scriptsize{$\pm$3.85}
    & \textbf{80.00\scriptsize{$\pm$0.00}}
    & 78.89\scriptsize{$\pm$5.09}
    & \textbf{20.67\scriptsize{$\pm$1.10}}
    & 18.12\scriptsize{$\pm$2.02}
    & 14.10\scriptsize{$\pm$2.56}
    \\

    {MAS-Zero}
    & 48.88\scriptsize{$\pm$5.08}
    & 47.78\scriptsize{$\pm$1.92}
    & \textbf{48.89\scriptsize{$\pm$1.92}}
    & 42.22\scriptsize{$\pm$5.09}
    & \textbf{45.56\scriptsize{$\pm$10.18}}
    & 42.22\scriptsize{$\pm$3.85}
    & 32.22\scriptsize{$\pm$5.09}
    & 33.33\scriptsize{$\pm$3.34}
    & 17.28\scriptsize{$\pm$1.39}
    & \textbf{17.36\scriptsize{$\pm$0.97}}
    & 15.07\scriptsize{$\pm$0.54}
    \\
     
    \bottomrule[0.12em]
    \end{tabular}
 }
\end{center}
\label{tab:main-context}
\vspace{-1em}
\end{table*}

This section expands on the empirical insights derived from experiments conducted along the four dimensions introduced in \Secref{sec:bg}.
\Secref{sec:verification-paradigm} examines different \emph{verification types}, comparing generative and scoring-based verifiers and analyzing the impact of judge specialization.
\Secref{sec:granularity} studies \emph{verification granularity}, characterizing each MAS’s preference between agent-level and iteration-level process verification for maximizing performance gains.
\Secref{sec:context} investigates \emph{context management strategies}, quantifying the trade-off between information retention and verification reliability.
Finally, \Secref{sec:analyse} decomposes performance by \emph{problem solvability}, isolating the role of process verification in recovering previously unsolvable queries versus improving stability across varying difficulty levels.

\subsection{Verification Paradigms}
\label{sec:verification-paradigm}
We analyze the extent of judge verification through different choices of verifiers. The experiments on \Tabref{tab:main} and \ref{tab:main-coding-science} are conducted by fixing the context to \textit{Raw/Full History} to establish a baseline.

\textbf{Intrinsic Difficulty of Partial Trajectory Verification.} As evidenced by the high variance in the \textit{Debate}, \textit{AFlow}, and \textit{MAS-Zero} rows of \Tabref{tab:main} and \ref{tab:main-coding-science}, process verification does not consistently yield monotonic performance improvements.  Although other methods show a natural sign of performance gains compared to the baseline, the pattern is noisy which highlights the challenge of evaluating the partial trajectories of MAS. 

\textbf{Generative Judges Outperform Scoring-based Reward Models.} Despite the high variance, a clear hierarchy is noticed when comparing the \textit{LLM-as-a-Judge} column against the \textit{Reward Model} (Skywork-Reward-V2) and \textit{Process Reward Model} (Qwen2.5-Math-PRM) columns. Generative judges achieved higher accuracy in 24 out of 36 experimental configurations across Agent-Call and Iteration levels. We attribute this performance gap to the nature of the training data for the reward model baselines. While the scalar baselines are highly specialized for specific domains (e.g., math steps or final outcomes), general-purpose generative judges appear better equipped to interpret the heterogeneous, out-of-distribution dynamics of MAS trajectories. This supports our hypothesis that the flexible, natural language reasoning of general LLMs offers a stronger steering signal for multi-agent systems than the rigid scalar values of specialized reward models.

\textbf{Diminishing Returns on Judge Scaling.} 
In \Tabref{tab:main-judge}, here, we fix the process verification level to \textit{Agent Call} (for granular viewing) and isolate the impact of the verifier's model architecture. The performance gap between using a smaller generalist judge (GPT-4o-mini) and a reasoning judge (GPT-5-mini) is significantly smaller than the gap observed when these models act as solvers. This indicates that the reasoning threshold required to \textit{verify} a multi-agent trajectory is lower than the threshold required to \textit{generate} it, suggesting that cost-effective, smaller models can serve as effective supervisors for larger systems.

\textbf{Understanding Specialized Verification Models.} Notably, the task-specific evaluator \textbf{FARE-20B} generally outperforms both general-purpose models (GPT-5-mini and GPT-4o-mini). For example, FARE-20B leads to the highest performance for \emph{MAS-Zero} on AIME24 (51.11$\pm$1.92) and GAIA (21.15$\pm$2.72), \emph{ADAS} on AIME25 (66.66$\pm$4.15) and GAIA (22.50$\pm$4.85) across all our experiments. These results suggest that evaluation-specific finetuning can yield additional gains over general-purpose LLMs-as-Judges.

\subsection{Verification Granularity}
\label{sec:granularity}
\Tabref{tab:main} shows that \textbf{no single verification granularity universally dominates across multi-agent systems}. Neither agent-level nor iteration-level verification consistently outperforms the other across MAS architectures, benchmarks, and verifier types, indicating that granularity choice is highly architecture-dependent rather than a global heuristic.

Nevertheless, \textbf{individual MAS often exhibit clear preferences}. \emph{Debate} and \emph{MAS-Zero} benefit more consistently from \emph{iteration-level} verification—especially with LLM-as-a-Judge—while \emph{AFlow} and \emph{ADAS} show stronger gains under \emph{agent-level} verification, suggesting better alignment with finer-grained control. \emph{MaAS} improves under both granularities but with benchmark-specific differences, whereas adaptive architectures such as \emph{DyLAN} display mixed behavior, favoring iteration-level verification for mathematical reasoning and agent-level verification for GAIA.

Overall, these results highlight that verification granularity interacts closely with MAS coordination dynamics: although no granularity is universally optimal, \textbf{aligning granularity with the MAS architecture is crucial for realizing the benefits of process-level verification.}

\subsection{Context Management Strategies}
\label{sec:context}
In \Tabref{tab:main-context}, we investigate the impact of information fidelity on the judge decision-making process. The process verification level is again fixed to \textit{Agent Call}.

\textbf{Context-Length vs Performance Trade-Off.} Results in \Tabref{tab:main-context} demonstrate that process verification with minimal context (Step-Only) performs comparably to, and occasionally better than, evaluation with Full Context. This suggests a strong need to understand the sweet-spot of context length that provides the right amount of context alongside not being vague. While full history contains necessary dependencies, it also introduces noisy tokens that can possibly degrade the judge's attention. Our data confirms a distinct context-length-performance trade-off in MAS process verification, where the utility of additional context diminishes as the noise-to-signal ratio increases.

\textbf{Summarized Context works best for MAS with long context.} The \textbf{MaAS} and \textbf{DyLAN} baselines produce MAS instances that take up significantly more tokens than other baselines, as shown in~\Tabref{tab:token_costs}. As a result, when evaluating these frameworks, the input context of the verifier must be carefully curated. In \Figref{fig:token_acc}, \textit{Summarized Context} strategies (Square/Triangle markers) consistently dominates \textit{Raw History} (Diamond markers) in both performance and efficiency. For frameworks like MaAS, the Raw strategy incurs a $\sim$3x token overhead while yielding comparatively lower accuracy. Further analysis/plots with respect to token growth, efficiency are presented in \appref{app:more-context}.

\begin{table}[t]
    \centering
    \caption{\textbf{Cost Efficiency and Trajectory Analysis.} Comparison of average token consumption per question across context strategies, alongside the average number of reasoning steps per trajectory. }
    \label{tab:token_costs}
    \vspace{-0.5em}
    \resizebox{\linewidth}{!}{
    \begin{tabular}{l c r r r r} 
        \toprule
        & \textbf{Avg.} & \multicolumn{4}{c}{\textbf{Avg. Tokens per Question by Context Strategy}} \\
        \cmidrule(l){3-6}
        \textbf{Architecture} & \textbf{Steps} & \textbf{Step-Only} & \textbf{Raw} & \textbf{Summary} & \textbf{Summ. + Step} \\
        \midrule
        DyLAN    & 4.4  & 1,196.6 & 2,280.7 & 467.1 & 1,650.2 \\
        MaAS     & 4.8  & 3,000.6 & 8,826.4 & 650.6 & 3,638.6 \\
        AFlow    & 5.2  & 365.8   & 1,754.9 & 574.1 & 913.7 \\
        ADAS     & 11.1 & 312.6   & 915.1   & 560.1 & 832.5 \\
        MAS-Zero & 21.6 & 330.8   & 1,409.7 & 589.9 & 861.0 \\
        \bottomrule
    \end{tabular}
    }
    \vspace{-0.5em}
\end{table}

\begin{figure}[ht!]
  \centering    
    \includegraphics[width=0.48\textwidth]{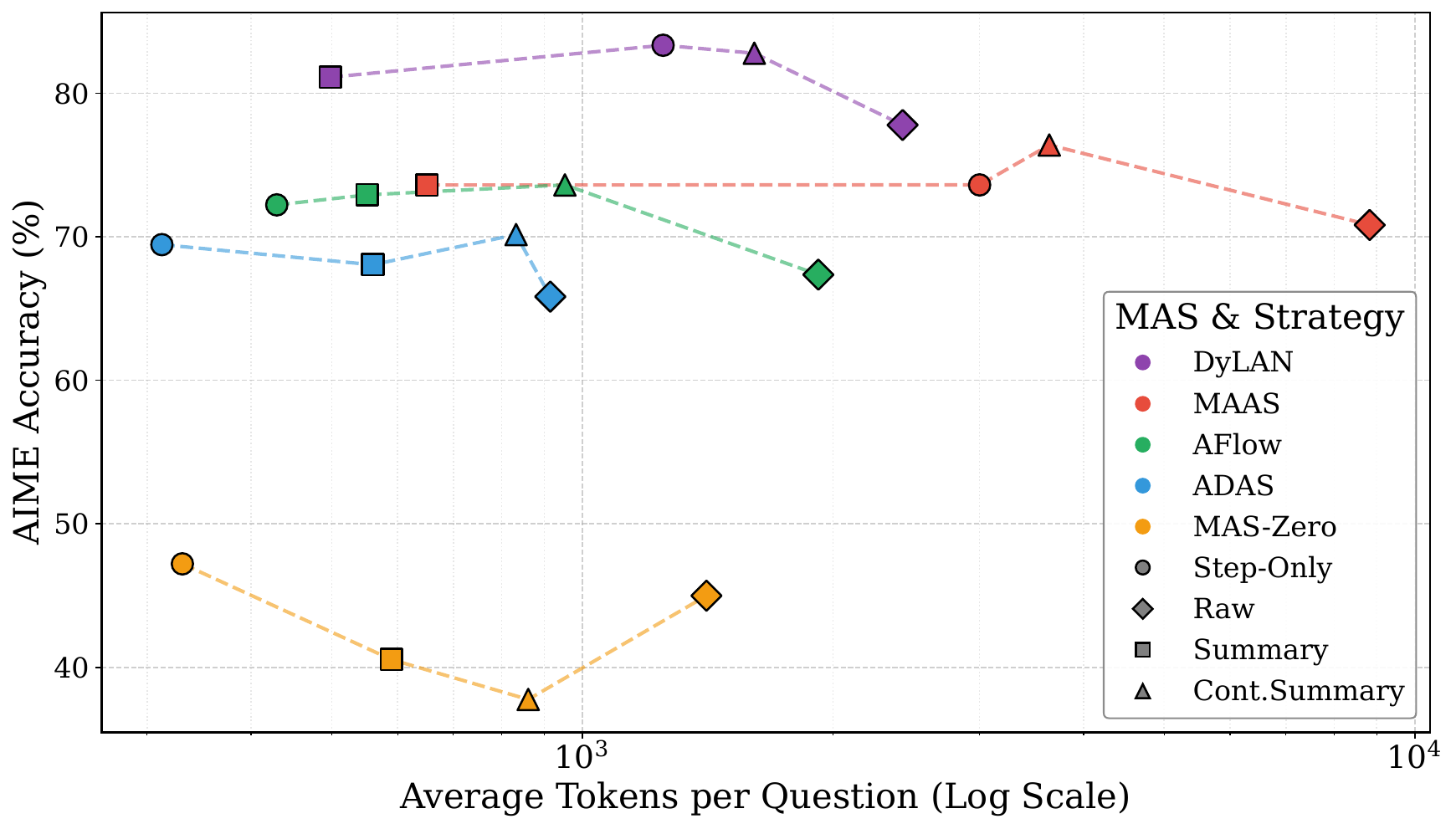}
    \caption{ Accuracy vs. Token Cost (Log Scale) across MAS architectures. Strategies closer to the top-left corner are to be preferred. Summarized contexts consistently outperform Raw History while consuming significantly fewer tokens.}
    \label{fig:token_acc}
\end{figure}

\noindent\textbf{Simply changing the judge or the context management can yield large gains.} Here, we aggregate our findings from \Tabref{tab:main-judge} and \Tabref{tab:main-context}, showing that even optimizing one of the choice of judge model and the choice of judge context management can yield substantial gains over blindly picking an evaluator. In~\Figref{fig:variance}, we plot the best performance of each MAS framework under different judge choices and context-management strategies on AIME24 and AIME25, denoted as ``Best Configuration''. We additionally compute \textbf{Pass@3} of standard MAS baseline runs. We find that some MAS frameworks outperform Pass@3 when paired with appropriately chosen verifications, such as AFLow on AIME24 or DyLAN on AIME25. Notably, the best judge configuration typically outperforms RM and PRM baselines by significant margins.

{In addition to these core dimensions, we investigate the underlying failure modes that contribute to occasional performance degradation. Specifically, we isolate instances of verifier misranking and demonstrate how the generative judge's reliability is highly sensitive to the provided context format. A detailed ablation study exploring this context-driven brittleness is provided in \appref{app:failure_modes}.}
\begin{figure*}[]
  \centering    
  \vspace{-0em}
    \includegraphics[width=\linewidth]{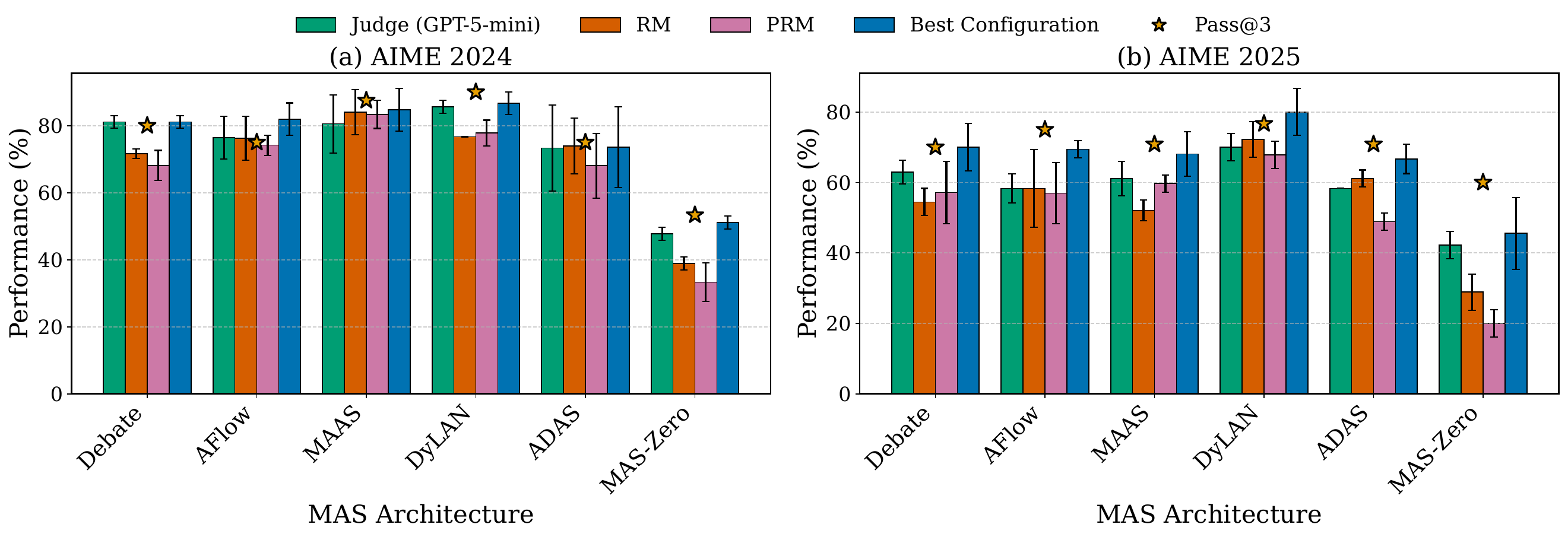}
    \caption{{Performance across MAS architectures} on Mathematical Reasoning Benchmarks using different process verification methods: Judge (GPT-5-mini), RM (Skywork-Reward-V2-Llama-3.1-8B), PRM (Qwen2.5-Math-PRM-7B), and Best Configuration. Error bars show standard deviation over three runs. Stars indicate Pass@3 performance.}
    \label{fig:variance}
    \vspace{-1.5em}
\end{figure*}

\subsection{Analysis of Solvability}
\label{sec:analyse}

We establish a difficulty baseline using a standard Chain-of-Thought (CoT) policy. We execute the base model (GPT-5-mini) on every question in the AIME 24 and AIME 25 datasets for $N{=}30$ independent trials to account for stochastic variance. The detailed stratification on both the datasets can be viewed in \appref{app:implementation}.

Building on this difficulty stratification, we proceed with a granular, question-wise comparative analysis to disentangle the impact of process supervision from inherent task complexity.

\textbf{Methodology.}\quad We conduct comparisons between the standard MAS baseline runs, the Agent call search executions guided by the \textbf{GPT-5-mini Judge}, and the best result acheiving configuration across our experiments. This formulation allows us to quantify these three critical metrics:
\begin{itemize}[nosep,leftmargin=18pt]
    \item \textbf{Performance Gain:} The absolute delta in pass rate between the baseline and the process-evaluated agents within each difficulty stratum.
    \item \textbf{Evaluator Stability:} The consistency of the judge's decisions, measured by the variance of outcomes (e.g., 0/3, 1/3, 2/3 or 3/3 success) across trials.
    \item \textbf{Intra-MAS Resurrection:} The recovery rate of baseline-failed questions (0/3), measuring whether process verification can salvage completely failed cases into partial or full successes.
    \end{itemize}

By isolating these variables, we can determine whether Process verification acts as a deterministic guide or introduces productive stochasticity. The resulting performance dynamics and stability profiles yields the following observations.

\noindent
\textbf{Process verification improves performance primarily by enhancing stability, with marginal utility on hard tasks.} As illustrated in \Figref{fig:gains}, the relative performance lift provided by our framework correlates positively with task difficulty. To pinpoint the source of this improvement, we analyze whether verification \textit{stabilizes} noisy baseline outputs or \textit{resurrects} fundamentally unsolvable problems. We observe that absolute gains remain marginal across all strata: \textit{Easy} (70--100\% baseline pass rate), \textit{Medium} (30--70\%), and \textit{Hard} (0--30\%). While ceiling effects likely constrain the Easy and Medium strata, the minimal shift in Hard tasks points to intrinsic solvability bounds within the underlying MAS. Instead, the primary driver of performance appears to be variance reduction; as shown in \Figref{fig:stability}, optimal configurations yield a marked increase in \textit{Stable Successes} (Green). This suggests that the utility of process verification lies less in solving novel hard problems, and more in systematically stabilizing solvable queries that suffer from generation variance.

\begin{figure}[ht!]
  \centering    
    \includegraphics[width=1\linewidth]{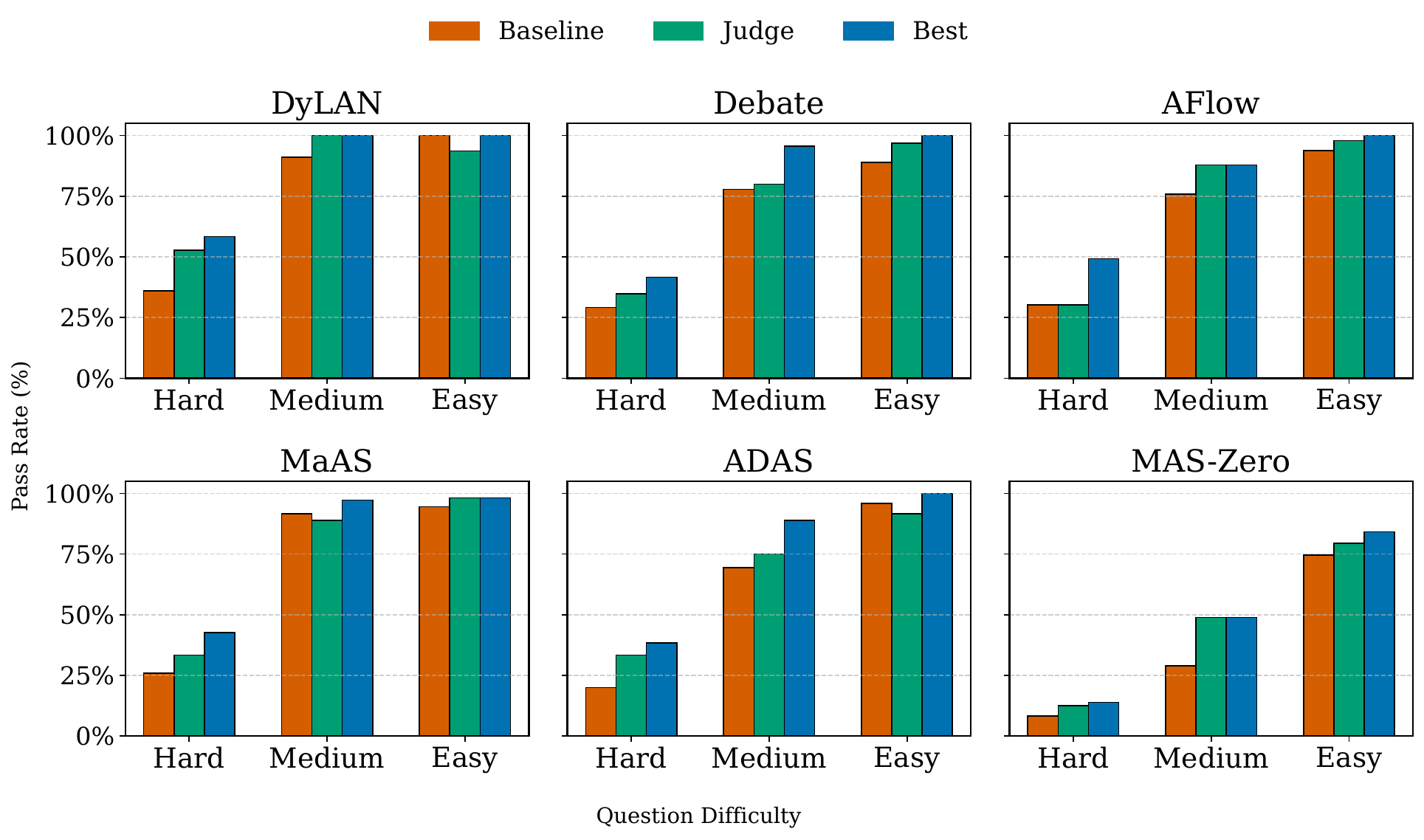}
    \caption{Average Pass rate on Mathematical Reasoning tasks by difficulty stratum. Baseline (red), Process Eval (green), and Best (blue) configurations compared across Hard, Medium, and Easy questions for six MAS architectures.}
    \label{fig:gains}
\end{figure}
\noindent
\textbf{Process verification encounters a fundamental solvability ceiling.} Complementing our findings on stability, analysis suggests that the framework's utility is strictly bounded by the agent's intrinsic generation horizon. As detailed in \Figref{fig:res}, for queries classified as ``fundamentally unsolvable'' (where the baseline policy yields 0\% success across all sampled trajectories), process verification rarely triggers a \textit{Full Resurrection} (Green). The majority of these instances result in \textit{Partial Resurrection} (Orange) or remain \textit{Still Failed} (Red). This confirms that while the verifier effectively prunes incorrect reasoning paths, it cannot completely synthesize correct ones that lie outside the agent's underlying reasoning manifold.

{To separate problems that are intrinsically unsolvable by the MAS generator from a failure of the verifier's ranking capacity, we conducted an exhaustive search experiment (Pass@128) across AIME questions of varying baseline difficulty. Our findings confirm that \textbf{for out-of-distribution queries, the MAS never generates the solution,} a fundamental generation limit that no verifier can overcome. Conversely, when the correct trajectory exists within the search space, the Generative Judge successfully identifies it and guides the search away from suboptimal solution trajectories. A detailed breakdown of this Pass@128 analysis is provided in \appref{app:solvability_ceiling}.}

\begin{figure}[ht!]
  \centering    
    \includegraphics[width=1\linewidth]{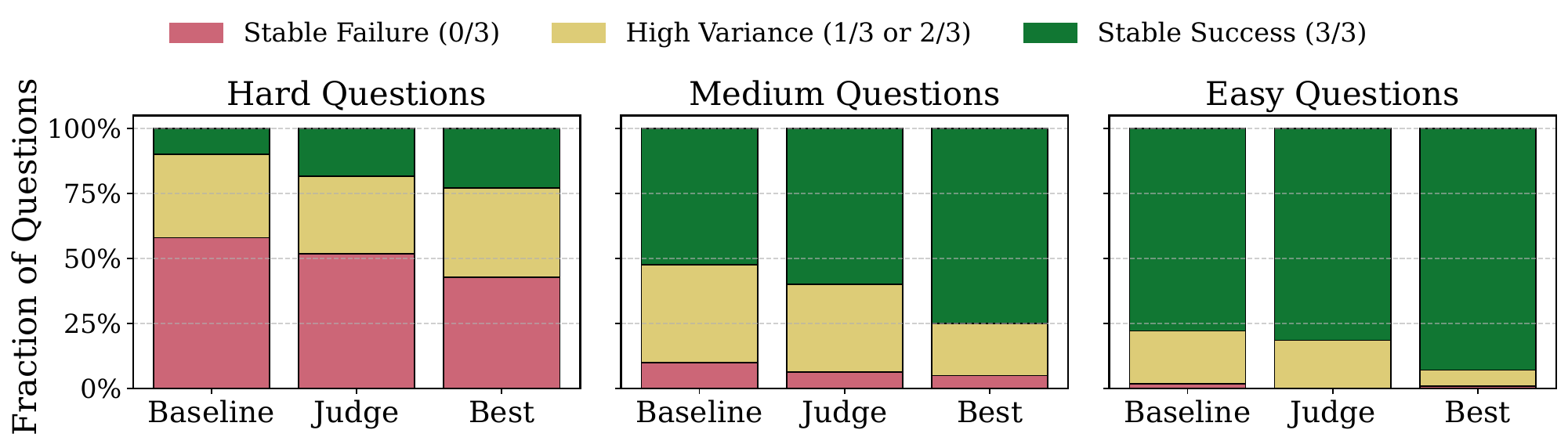}
    \caption{Stability distribution aggregated across all MAS architectures on mathematical reasoning tasks. Stacked bars show trial outcome proportions by difficulty: Stable Failure (red, 0/3), High Variance (yellow, 1-2/3), and Stable Success (green, 3/3) for Baseline, Judge, and Best configurations.}
    \label{fig:stability}
\end{figure}

\begin{figure}[ht!]
  \centering    
    \includegraphics[width=1\linewidth]{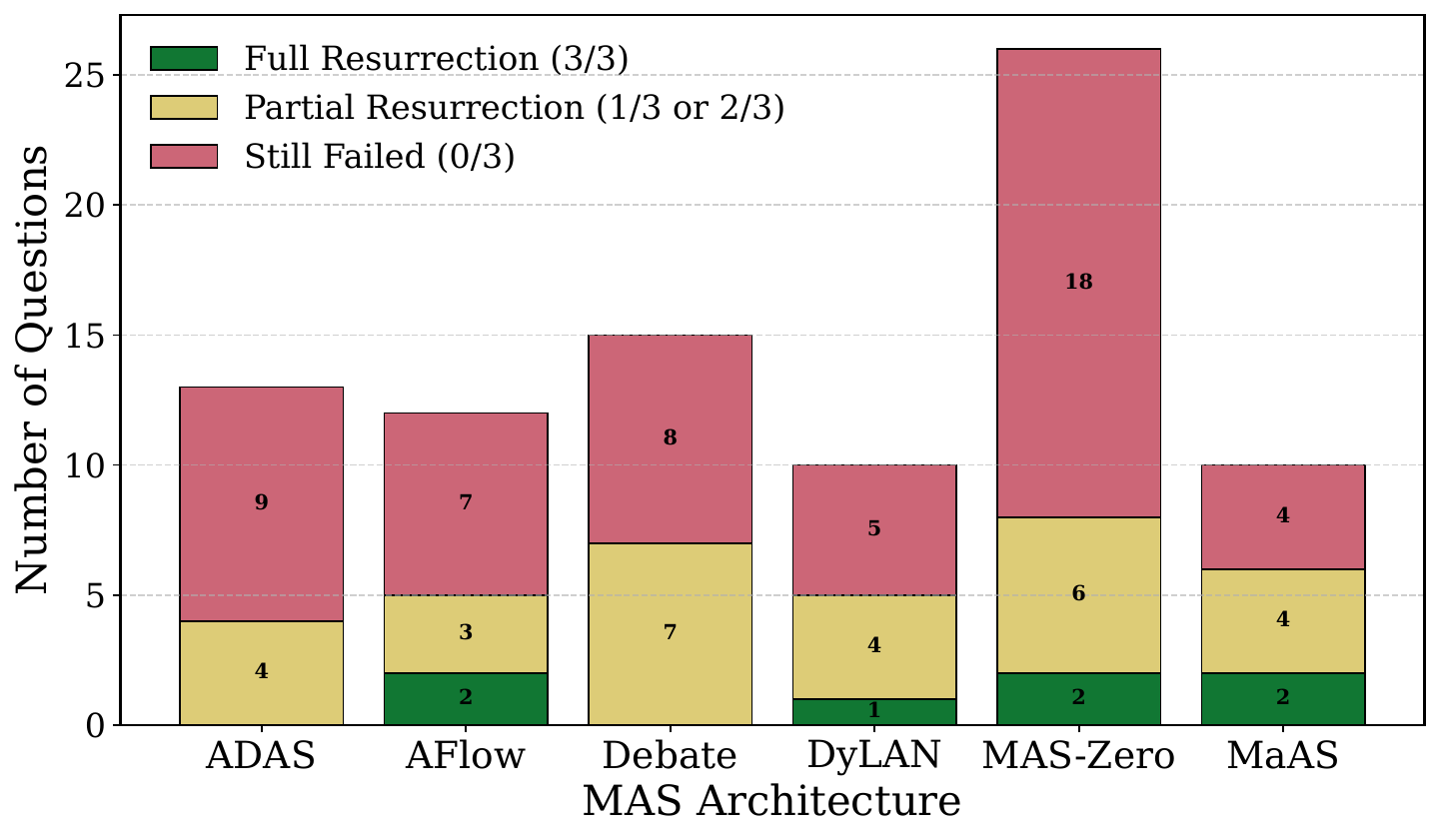}
    \caption{\textbf{Intra-MAS Resurrection Analysis.} The chart visualizes the recovery yield of the best process verification configuration from our experiments on these specific failure cases. \textit{Full Resurrection} (Green) demonstrates that process-guided search can rarely navigate to solution states that are unreachable by the baseline.}
    \label{fig:res}
    \vspace{-1em}
\end{figure}

\section{Related Work}
\label{sec:related}

\textbf{Inference-time Automatic MAS.}\quad
Recent advancements in Multi-Agent Systems have shifted towards inference-time optimization, where agent structures or workflows are adapted dynamically to the input query. Frameworks like \textbf{MASS}~\citep{zhou2025multiagentdesignoptimizingagents} and \textbf{MaAS}~\citep{zhang2025multiagent} utilize rejection sampling and masking mechanisms to select effective sub-agent configurations on the fly. 
More aggressive optimization approaches, such as \textbf{ADAS}~\citep{hu2025automated} and \textbf{AFlow}~\citep{zhang2025aflow}, treat MAS execution as a code generation task, employing search algorithms like MCTS to discover optimal workflows. 
\textbf{MAS-Zero}~\citep{Ke2025MASZero}, being a noble outlier, attempts fully inference-time adaptation without external validation sets. With the existing architectures focus being largely on overall architectural optimization, our protocol effectively adds a quality-assurance layer on top of these inference-time frameworks. This complements with the current landscape, by steering the intermediate outputs of the MAS.

\textbf{Verification for MAS.}\quad
While early strategies for verification relied on scalar RMs to guide search, recent work suggests that generative verifiers may offer superior guidance for complex reasoning. ~\citet{zhang2025generativeverifiersrewardmodeling} demonstrate that trained generative verifiers outperform scalar RMs in Best-of-$N$ reranking tasks.
Verification serves as the engine for Test-Time Scaling (TTS). ~\citet{snell2024scaling} and ~\citet{liu20251bllmsurpass405b} show that compute-optimal scaling strategies rely heavily on the interaction between the policy model and the process verifier. ~\citet{zhou2025evaluating} further introduced the JETTS benchmark to systematically evaluate these judges, showing that generative evaluators can match or exceed the performance of specialized reward models in reranking contexts. Our work extends this by investigating how these scaling benefits translate to the granular, intermediate steps of Multi-Agent Systems.

A critical factor in process verification is the alignment between the generator and the verifier. ~\citet{chen2025llmevaluatorspreferreason} observe a strong positive correlation between generation capability and evaluation accuracy when the same model is used for both roles, a setup we adopt by using GPT-5-mini as both for the generating agent and judge. However, this setup introduces risks of self-preference bias~\citep{panickssery2024llmevaluatorsrecognizefavor}, where evaluators may favor their own reasoning patterns. Additionally,~\citet{krumdick2025freelabelslimitationsllmasajudge} find that evaluator performance degrades significantly when the task difficulty exceeds the model's latent reasoning capability, a limitation we analyze through our Resurrection results.

\section{Conclusion}
\label{sec:conclusion}
In this work, we conduct a comprehensive empirical study of process verification in multi-agent systems, systematically examining the interactions among verification type, verification granularity, and context management strategies. Our results show that, although process-level verification can improve stability on queries that are already within the agents’ solvable regime, it rarely enables recovery from fundamentally unsolvable states, underscoring the limits of current verification-based steering mechanisms.
We further release a modular engineering framework that serves as a plug-and-play wrapper for \emph{any} off-the-shelf MAS and \emph{any} process verifier, enabling reproducible evaluation and future extensions. Overall, our findings suggest that integrating MAS with automated process verifiers is \emph{\textbf{not yet}} a universal solution: reliably assessing and ranking partial multi-agent trajectories remains a central open challenge that cannot be resolved by straightforward verification alone.

\section*{Impact Statement}
This research advances methods for making large language model collaboration more capable, reliable, and interpretable, through a systematic empirical study of how process verification interacts with the existing multi-agent sysmtems. There are many potential societal consequences of our work, none of which we feel must be specifically highlighted here.

\section*{Acknowledgement}
\label{sec:acknowledgement}
We thank all reviewers, AC, and SAC for their valuable comments. HW is supported by Amazon Faculty Research Award, Microsoft AI \& Society Fellowship, NSF CAREER Award IIS-2340125, NIH grant R01CA297832, and NSF grant IIS-2127918.

{
\bibliography{ref,mas_r1,mas_zero}
\bibliographystyle{icml2026}
}

\onecolumn
\appendix
\clearpage
\section*{\LARGE Appendix}
\markboth{Appendix}{Appendix}

In \appref{app:implementation}, we provide our implementation details of the experiments, including: 
\begin{itemize}[nosep]
    \item \textbf{dataset details}~(\appref{app:datasets}),
    \item \textbf{baseline details}~(\appref{app:baselines}),
    \item \textbf{prompt templates} used in LLM reasoning~(\appref{app:prompts}),
    \item and \textbf{code implementation of \ours}~(\appref{app:code-template}).
\end{itemize}
In \appref{app:more}, we present additional empirical results, including:
\begin{itemize}[nosep]
    \item \textbf{extended experiments of MAS-Zero}~(\appref{app:more-mas-zero}),
    \item additional \textbf{benchmark analysis}~(\appref{app:more-benchmark})
    \item additional \textbf{analysis on context}~(\appref{app:more-context}), and
    \item {\textbf{case study}}~(\appref{app:more-case-study}).
\end{itemize}

\section{Implementation Details}
\label{app:implementation}

\subsection{Datasets}
\label{app:datasets}
We evaluate our framework on three distinct benchmarks, namely AIME24, AIME25 and GAIA. Both AIME24 and AIME25 comprise of 30 questions each with the information extraction subset of GAIA containing 103 questions.

\textbf{AIME 24 \& 25 (Mathematical Reasoning).}\\
For the AIME datasets, we define difficulty empirically. We run a standard Chain-of-Thought (CoT) baseline 30 times per question. Questions are then stratified into \textit{Easy}, \textit{Medium}, and \textit{Hard} strata based on the baseline's success rate, as detailed in \Tabref{tab:difficulty_stats}. This ensures that our ``Hard'' category represents tasks that are stochastically unstable or consistently failing under standard inference.
\begin{table}[h]
    \centering
    \caption{\textbf{Benchmark Difficulty Distribution.}}
    \label{tab:difficulty_stats}
    \resizebox{0.5\linewidth}{!}{
    \setlength{\tabcolsep}{6pt}
    \begin{tabular}{l c c c c}
        \toprule
        \textbf{Dataset} & \textbf{Avg. Pass Rate} & \textbf{Easy} & \textbf{Medium} & \textbf{Hard} \\
         & (CoT Baseline) & \small{($>$20/30)} & \small{(11--20/30)} & \small{(0--10/30)} \\
        \midrule
        \textbf{AIME 24} & 58.3\% & 13 \small{(43.3\%)} & 9 \small{(30.0\%)} & 8 \small{(26.7\%)} \\
        \textbf{AIME 25} & 42.0\% & 10 \small{(33.3\%)} & 4 \small{(13.3\%)} & 16 \small{(53.3\%)} \\
        \bottomrule
    \end{tabular}
    }
\end{table}

\textbf{GAIA (Information Extraction).} \\
For the General AI Assistant benchmark (GAIA), we utilize the validation set of the \textit{Information Extraction} subset ($N=103$). Unlike AIME, GAIA difficulty is intrinsic, pre-annotated by the benchmark creators into Levels 1--3 based on the number of steps and tools required. As shown in \Tabref{tab:gaia_stats}, this dataset is heavily skewed towards multi-step tool use, with 90.3\% of tasks requiring external tool calls (e.g., web search, file processing), making it an ideal testbed for evaluating process verification in tool-augmented settings.

\begin{table}[h]
    \centering
    \caption{\textbf{GAIA Statistics.}}
    \label{tab:gaia_stats}
    \resizebox{0.5\linewidth}{!}{
    \setlength{\tabcolsep}{12pt}
    \begin{tabular}{l c c c c}
        \toprule
        \textbf{Metric} & \textbf{Level 1} & \textbf{Level 2} & \textbf{Level 3} & \textbf{Total} \\
        \midrule
        \textbf{Count} & 39 & 52 & 12 & 103 \\
        \midrule
        \multicolumn{5}{l}{\textit{\textbf{Tool Requirements:}}} \\
        \multicolumn{5}{l}{\quad $\bullet$ \textbf{Tool Required:} 93 tasks (90.3\%)} \\
        \multicolumn{5}{l}{\quad $\bullet$ \textbf{No Tools:} 10 tasks (9.7\%)} \\
        \bottomrule
    \end{tabular}
    }
\end{table}

\subsection{Baselines}
\label{app:baselines}
\Tabref{tab:baselines_config} details the specific experimental protocols for each baseline. 
\begin{itemize}[nosep]

\item \textbf{Debate}~\citep{du2024improving}: We adopt the official LLM-Debate code by using the unified CoT prompt for initial thought generation, as in this paper, and also explicitly instructing each agent to refine its response based on the peer agent responses. We run two agents for three iterations of debate, and for fair comparison, randomly select one of the final-round answers as the output.

\item \textbf{DyLAN}~\citep{liu2024dynamicllmpowered}: The Dynamic LLM-Powered Agent Network is re-implemented for Mathematical reasoning tasks by adopting from the MATH benchmark prompts. Tool calling is enabled with the usage of the DuckduckGo websearch call while dealing with GAIA benchmark with the roles of the agents adopted from MMLU implementation in the existing codebase.

\item \textbf{AFlow}~\citep{zhang2025aflow} and  \textbf{MaAS}~\citep{zhang2025multiagent}: For mathematical reasoning benchmarks, we retained the operator set and prompt libraries originally designed for the MATH benchmark, applying only minimal schema alignments to handle the distinct output formats of AIME. For the \textbf{GAIA} benchmark, we significantly re-engineered the agent space: \begin{itemize} \item \textbf{Tool Integration:} We integrated a \textbf{DuckDuckGo} search tool directly into the execution environment, enabling agents to query external information required for open-ended assistant tasks. \item \textbf{Prompt Adaptation:} We adopted the evaluation prompt templates from the official GAIA implementation to ensure consistent instruction adherence. \item \textbf{Operator Pruning:} For the GAIA Information Extraction subset, we excluded the \texttt{Programmer} operator from the search space, focusing the architecture on retrieval and linguistic synthesis rather than code execution. \end{itemize}

\item \textbf{MAS-Zero}~\citep{Ke2025MASZero}: As the original implementation of MAS-Zero natively supports \textbf{AIME 24}, we extended this configuration to \textbf{AIME 25} without modification, utilizing the same prompt structure and atomic baselines. For \textbf{GAIA}, we expanded the architecture's fundamental building blocks. In addition to the standard four atomic baselines (e.g., CoT, Reflexion), we introduced a specialized \textbf{WebSearch Agent} primitive. This new block combines a DuckDuckGo retrieval interface with a Chain-of-Thought (CoT) wrapper, allowing the generated code to not only fetch results but also reason over the retrieved context before integrating it into the solution path.

\item \textbf{ADAS}~\citep{hu2025automated}: Similar to MAS-Zero, the ADAS implementation required prompt refinement to handle AIME24 and AIME25. For \textbf{GAIA}, we applied the same architectural augmentation as described for MAS-Zero: the meta-agent was provided access to the \textbf{WebSearch Agent} primitive, effectively expanding the search space to include tool-augmented reasoning policies suitable for real-world information extraction.

\end{itemize}
For optimization-centric frameworks like AFlow, MaAS, and ADAS, we employ a strict validation/test split. Additionally, for generative architectures (MAS-Zero, ADAS), we restrict the primary evaluation to minimal iteration settings (1 and 5 iterations, respectively) and omit fixed baseline blocks (e.g., CoT, CoT-SC, Reflexion, etc. ) during evaluation. This design choice isolates the performance of the generated agents and manages the exponential token growth inherent to these methods (see \Tabref{tab:token_costs}). Extended scalability results for MAS-Zero are provided in \appref{app:more}.

\begin{table}[h]
    \centering
    \caption{\textbf{Summary of MAS Baseline Configurations.} We distinguish between architectures evaluated on the full dataset versus those requiring optimization splits. Iteration limits for ADAS and MAS-Zero are imposed to balance computational feasibility with search depth.}
    \label{tab:baselines_config}
    \resizebox{0.8\linewidth}{!}{
    \setlength{\tabcolsep}{14pt}
    \begin{tabular}{l l l l}
        \toprule
        \textbf{Baseline} & \textbf{Protocol} & \textbf{Configuration Details} \\
        \midrule
        \textbf{Debate} & Full Dataset & 2 Agents, 3 Rounds \\
        \textbf{DyLAN} & Full Dataset & 4 Agents, $T=4$ Rounds \\
        \textbf{AFlow} & Val/Test Split & 20 Optimization Rounds, 2 Validation Rounds \\
        \textbf{MaAS} & Train/Test Split & Sample Size $N=4$  \\
        \textbf{ADAS} & Val/Test Split & 5 Iterations. Additional WebSearch Baseline for GAIA \\
        \textbf{MAS-Zero} & Full Dataset & 1 Iteration. Additional WebSearch Baseline for GAIA \\
        \bottomrule
    \end{tabular}
    }
\end{table}

\subsection{Prompt Templates}
\label{app:prompts}
This section presents the prompt templates for the various judges used in our verification framework.

\begin{tcolorbox}[
    breakable,
    enhanced,
    left=-0.5cm, right=-0.5cm, top=2pt, bottom=2pt,
    enlarge top by=0.1cm, 
    enlarge bottom by=0.1cm, 
    title={\hspace{1cm} LLM-as-a-Judge}, 
    fonttitle=\bfseries\small
]
\begin{quote}
Please act as an impartial judge and evaluate the quality of multiple responses provided by AI assistants to the user prompt displayed below. You will be given several candidate responses.

Your task is to rank these responses from best to worst based on their correctness, reasoning quality, and completeness. Consider:
- Correctness of the final answer
- Quality and clarity of reasoning
- Completeness of the solution
- Logical consistency

Be as objective as possible. Avoid any biases such as length or stylistic elements like formatting.

Before providing your ranking, think through the evaluation process and output your thoughts as an explanation.

After providing your explanation, you must output the ranking as a comma-separated list of candidate numbers (e.g., "2,1,3" means candidate 2 is best, candidate 1 is second, candidate 3 is worst). Please enclose your ranking in <ranking> and </ranking> tags.
\end{quote}
\end{tcolorbox}

\begin{tcolorbox}[
    breakable,
    enhanced,
    left=-0.5cm, right=-0.5cm, top=2pt, bottom=2pt,
    enlarge top by=0.1cm, 
    enlarge bottom by=0.1cm, 
    title={\hspace{1cm} FARE-20B}, 
    fonttitle=\bfseries\small
]
\begin{quote}
PROMPT\_RANKING\_SYSTEM = """
Please act as an impartial judge and evaluate the quality of the responses provided by three AI assistants to the user prompt displayed below. You will be given assistant A's answer, assistant B's answer, and assistant C's answer. Your job is to determine which assistant's answer is better.
If assistant A is better, output [A]. If assistant B is better, output [B]. If assistant C is better, output [C].

Here are some rules for evaluation
(1) When evaluating the assistants' answers, identify any mistakes or inaccurate information. Focus on the content each response and select the response that is logically sound and error free.
(2) If both responses contain inaccurate information, select the response that arrives at the correct response
(3) Avoid any biases, such as order of responses, length, or stylistic elements like formatting

Before outputting your final judgment, provide an explanation of your judgment. Your explanation should discuss why your chosen response is better based on the evaluation criteria. The explanation should concretely discuss strengths and weaknesses of both answers.
After outputting your explanation, provide your final judgment. Use the following format:
Explanation: Your explanation here
Verdict: Your final verdict
""".strip()

PROMPT\_PAIRWISE = """
[User Question]
{instruction}

[The Start of Assistant A's Answer]
\{response\_a\}
[The End of Assistant A's Answer]

[The Start of Assistant B's Answer]
\{response\_b\}
[The End of Assistant B's Answer]

[The Start of Assistant C's Answer]
\{response\_c\}
[The End of Assistant C's Answer]
""".strip()
\end{quote}
\end{tcolorbox}

\begin{tcolorbox}[
    breakable,
    enhanced,
    left=-0.5cm, right=-0.5cm, top=2pt, bottom=2pt,
    enlarge top by=0.1cm, 
    enlarge bottom by=0.1cm, 
    title={\hspace{1cm} Summarizer Prompt for PRM/ RM/ FARE-20B/ GPT-5-mini}, 
    fonttitle=\bfseries\small
]
\begin{quote}
You are a summarization assistant for multi-agent system processes.

Your task is to summarize the given context into a concise and informative summary.
The total length of the summary should be fewer than \{max\_context\_length\} tokens.

IMPORTANT PRINCIPLES:
- This is NOT style normalization.
- This is NOT reformatting to a fixed template.
- The cleaned solution should look like the SAME author wrote it,
  just more concisely.

STRICT REQUIREMENTS:
1) Faithfulness:
   - Do NOT change the meaning.
   - Do NOT change the final answer(s).
2) Reasoning preservation:
   - Keep ALL logical reasoning steps and ALL necessary mathematical derivations.
   - You may remove repeated steps, detours, or restatements ONLY if the reasoning remains intact.
3) Language \& style preservation:
   - Preserve the original narrative tone (e.g. exploratory, explanatory, corrective).
   - Preserve structural choices used in the original solution
     (such as step labels, section headers, or paragraph structure),
     unless they are clearly redundant.
4) Formatting:
   - Do NOT convert narrative explanations into bullet fragments or terse notes.
   - Do NOT collapse structured explanations into formula-only derivations.
   - Keep LaTeX for mathematics.

Output ONLY the cleaned solution text. No extra commentary.
\end{quote}
\end{tcolorbox}

\begin{tcolorbox}[
    breakable,
    enhanced,
    left=-0.5cm, right=-0.5cm, top=2pt, bottom=2pt,
    enlarge top by=0.1cm, 
    enlarge bottom by=0.1cm, 
    title={\hspace{1cm} General prompt for GAIA}, 
    fonttitle=\bfseries\small
]
\begin{quote}
You are a general AI assistant with access to web search capabilities.
If you are asked for a number, don't use comma to write your number neither use units such as \$ or
percent sign unless specified otherwise.
If you are asked for a string, don't use articles, neither abbreviations (e.g. for cities), and write the
digits in plain text unless specified otherwise.
If you are asked for a comma separated list, apply the above rules depending of whether the element
to be put in the list is a number or a string.

QUESTION: \{question\}

If you need to look up information to answer this question, you can use the web\_search function available to you.
Call it with specific search queries when you need factual information, current data, or verification.

You can freely reason in your response, please: 
1. include the thinking process in your response, in a detailed and extended manner, and after that, 
2. enclose the final answer within <answer></answer> tags

\end{quote}
\end{tcolorbox}

\definecolor{codegreen}{rgb}{0,0.6,0}
\definecolor{codegray}{rgb}{0.5,0.5,0.5}
\definecolor{codepurple}{rgb}{0.58,0,0.82}
\definecolor{backcolour}{rgb}{0.95,0.95,0.92}

\lstdefinestyle{mystyle}{
    commentstyle=\color{codegreen},
    keywordstyle=\color{blue},
    numberstyle=\tiny\color{codegray},
    stringstyle=\color{codepurple},
    basicstyle=\ttfamily\footnotesize,
    breakatwhitespace=false,         
    breaklines=true,                 
    captionpos=b,                    
    keepspaces=true,                 
    showspaces=false,                
    showstringspaces=false,
    showtabs=false,                  
    tabsize=2,
    language=Python
}

\subsection{Code Templates}
\label{app:code-template}
Here we will define the code templates that serve the primary focus of our process verification framework. We will go over the parallel search decorator that acts as the middleware between the client(MAS) and the Server(Judge). We also show the MASBase main class which helps in managing the context states and also serve as an independent handler of the flow of process verification.

\begin{tcolorbox}[
    breakable,
    enhanced,
    left=0.2cm, right=0.2cm, top=2pt, bottom=2pt,
    enlarge top by=0.1cm, 
    enlarge bottom by=0.1cm, 
    title={\hspace{1cm}Parallel Search Decorator}, 
    fonttitle=\bfseries\small,
    colback=white 
]
\begin{lstlisting}[style=mystyle]
def llm_parallel_search_decorator(llm_func):
    """
    Decorator to run the LLM call function MAX_PARALLEL_SEARCH_CALLS times in parallel,
    send the results to the server via client, and pick the best response.
    """
    async def gather_calls(*args, **kwargs):
        # Call the original LLM func in parallel
        tasks = [llm_func(*args, **kwargs) for _ in range(MAX_PARALLEL_SEARCH_CALLS)]
        responses = await asyncio.gather(*tasks)
        return responses

    def send_to_server(responses, client: BaseClient, task_type=None):
        # Sends all candidate responses to server (expects 'server_client' conforming to your client API)
        return client.send_request(task_type, "judge", responses)
    
    def decorator_wrapper(*args, **kwargs):
        """
        kwargs must include 'client', and optionally 'task_type'
        """
        task_type = kwargs.pop('task_type', None)
        client = kwargs.pop('client', None)
        if client is None:
            raise ValueError("A 'client' must be provided as a kwarg to the decorated function.")

        # Run the LLM calls in parallel & collect results
        responses = asyncio.run(gather_calls(*args, **kwargs))

        # Send all results to the server judge and get rankings
        server_result = send_to_server(responses, client, task_type)
        rankings = server_result.get("rankings", list(range(len(responses))))
        best_idx = rankings[0] if rankings else 0

        return responses[best_idx]

    return decorator_wrapper
\end{lstlisting}
\end{tcolorbox}

\begin{tcolorbox}[
    breakable,
    enhanced,
    left=0.2cm, right=0.2cm, top=2pt, bottom=2pt,
    enlarge top by=0.1cm, 
    enlarge bottom by=0.1cm, 
    title={\hspace{1cm}MASBase Class}, 
    fonttitle=\bfseries\small,
    colback=white 
]
\begin{lstlisting}[style=mystyle]
class MASBase:
    def __init__(self):
        self.trajectory = []
        self.trajectory_lock = asyncio.Lock()

    @property
    def progress(self):
        return json.dumps({f"step_{i+1}": step for i, step in enumerate(self.trajectory)}, indent=4)

    @staticmethod
    def update_trajectory(func):
        """
        Appends the latest step/result to the trajectory and updates the progress.
        Should be called automatically by wrappers of generation calls.
        """
        async def wrapper(self, *args, **kwargs):
            # Get the original result from the decorated function
            result = await func(self, *args, **kwargs)
            
            # Extract content for trajectory while preserving original result
            trajectory_content = result
            if isinstance(result, tuple) or isinstance(result, list):
                trajectory_content = result[0]
            # Convert dict to string for trajectory storage
            if isinstance(trajectory_content, dict):
                trajectory_content = json.dumps(trajectory_content)
            
            async with self.trajectory_lock:
                self.trajectory.append(trajectory_content)

            return result
        return wrapper

    def run(self): 
        raise NotImplementedError("Subclasses must implement this method")

\end{lstlisting}
\end{tcolorbox}

\begin{tcolorbox}[
    breakable,
    enhanced,
    left=0.2cm, right=0.2cm, top=2pt, bottom=2pt,
    enlarge top by=0.1cm, 
    enlarge bottom by=0.1cm, 
    title={\hspace{1cm} MASBase Adaptation on AFlow for Agent Call Process Verification }, 
    fonttitle=\bfseries\small,
    colback=white 
]
\begin{lstlisting}[style=mystyle]
class MASAFlow(MASBase):
    """
    MAS wrapper for AFlow workflows - analogous to MASDebate for debate.
    Wraps workflow execution with trajectory tracking and process verification.
    
    Usage (parallel to MASDebate):
        mas_aflow = MASAFlow(workflow, problem, expected_answer, task_type)
        result = await mas_aflow.run()
    """
    
    def __init__(self, workflow, task_type: str = "math"):
        super().__init__()
        self.workflow = workflow
        self.task_type = task_type
        self.problem = ""  # Will be set per call
        
        # Inject decorators into workflow operators (only once)
        self._inject_decorators()

        
    @MASBase.update_trajectory
    @llm_parallel_search_decorator
    async def _call_custom(self, *args, **kwargs):
        """Decorated custom operator call"""
        result = await self._original_custom(*args, **kwargs)
        return result
    
    @MASBase.update_trajectory
    @llm_parallel_search_decorator
    async def _call_programmer(self, *args, **kwargs):
        """Decorated programmer operator call"""
        return await self._original_programmer(*args, **kwargs)
    
    @MASBase.update_trajectory
    @llm_parallel_search_decorator
    async def _call_sc_ensemble(self, *args, **kwargs):
        """Decorated sc_ensemble operator call"""
        return await self._original_sc_ensemble(*args, **kwargs)
    
    @MASBase.update_trajectory
    @llm_parallel_search_decorator
    async def _call_answer_generate(self, *args, **kwargs):
        """Decorated answer_generate operator call"""
        return await self._original_answer_generate(*args, **kwargs)
    
    async def __call__(self, problem: str):
        """
        Execute the workflow with trajectory tracking for a specific problem.
        Returns same format as normal workflow: (final_answer, cost)
        """
        # Update problem for this execution
        self.problem = problem
        # Reset trajectory for new problem
        self.trajectory = []
        # Execute workflow
        result = await self.workflow(problem)
        return result
    
    async def run(self, problem: str):
        """Alias for __call__ for backwards compatibility"""
        return await self(problem)
\end{lstlisting}
\end{tcolorbox}

\begin{tcolorbox}[
    breakable,
    enhanced,
    left=0.2cm, right=0.2cm, top=2pt, bottom=2pt,
    enlarge top by=0.1cm, 
    enlarge bottom by=0.1cm, 
    title={\hspace{1cm} Iteration Level Process Verification on AFlow Workflow}, 
    fonttitle=\bfseries\small,
    colback=white 
]
\begin{lstlisting}[style=mystyle]
async def __call__(self, problem: str):
        """
        Wrapper that triggers the parallel search.
        The decorator runs the logic 3 times in parallel and the Judge picks the best.
        """
        # Call the decorated function with required Judge parameters
        best_result = await self._run_logic_async(
            problem=problem,
            question=problem,      # REQUIRED: User prompt for the Judge
            task_type="math"    # REQUIRED: Triggers the Math/QA prompt template
        )
        
        # Unpack the best result
        if isinstance(best_result, dict):
            return best_result["output"], best_result["cost"]
        else:
            return str(best_result), 0

    # -------------------------------------------------------------------------
    # 2. The Decorated Logic (Runs N times in Parallel)
    # -------------------------------------------------------------------------
    @llm_parallel_search_decorator
    async def _run_logic_async(self, problem: str, **kwargs):
        """
        The core workflow logic.
        Runs N times in parallel. Returns a full trace for the Judge to evaluate.
        """
        import re

        # --- Step 1: Generate Candidates ---
        # First candidate: concise final numeric answer
        instr1 = prompt_custom.CANDIDATE_PROMPT_1
        resp1 = await self.custom(input=problem, instruction=instr1)

        # Second candidate: step-by-step
        instr2 = prompt_custom.CANDIDATE_PROMPT_2
        resp2 = await self.custom(input=problem, instruction=instr2)

        # --- Step 2: Ensemble ---
        # Handle dynamic method name check from your snippet
        solutions = [resp1['response'], resp2['response']]
        if hasattr(self, 'sc_ensemble'):
            ensemble = await self.sc_ensemble(solutions=solutions, problem=problem)
        else:
            ensemble = await self.sc(solutions=solutions, problem=problem)
        
        chosen = ensemble['response']

        # --- Step 3: Extraction ---
        nums = re.findall(r'\b-?\d+\b', chosen)
        candidate_num = nums[-1] if nums else ''

        # --- Step 4: Programmer Verification ---
        verification_log = ""
        prog_verify = await self.programmer(problem=problem, analysis=f"Verify whether the final numeric answer is {candidate_num}. Return only the final integer if verified, otherwise return a different integer computed.")
        verified = prog_verify.get('output', '').strip()
        verification_log = f"Verification Output: {verified}"
        
        if verified and re.search(r'\b-?\d+\b', verified):
            final_answer = verified
        else:
            # Fallback: compute from scratch
            prog_full = await self.programmer(problem=problem, analysis="Compute and return only the final numeric answer (integer).")
            final_answer = prog_full.get('output', '').strip() or candidate_num
            verification_log += f"\nFallback Calculation: {final_answer}"

        # Get total cost
        total_cost = self.llm.get_usage_summary().get("total_cost", 0)

        # --- Construct Trace for the Judge ---
        # This is what allows the Judge to rank this attempt against the parallel others
        trace_str = (
            f"--- Candidate 1 (Concise) ---\n{resp1.get('response', '')}\n\n"
            f"--- Candidate 2 (Step-by-Step) ---\n{resp2.get('response', '')}\n\n"
            f"--- Ensemble Decision ---\n{chosen}\n\n"
            f"--- Programmer Verification ---\n{verification_log}\n\n"
            f"--- Final Answer ---\n{final_answer}"
        )

        # --- Return Dictionary for Decorator ---
        return {
            "response": trace_str,      # The reasoning trace (Primary for Judge)
            "output": final_answer,     
            "cost": total_cost          
        }
\end{lstlisting}
\end{tcolorbox}
Since all the Baselines used have their own prompt templates presented in their own paper, we will look to address the templates that included drastic changes from the original which we had to implement such as the WebSearch agent for ADAS and MAS-Zero.
\label{app:code}

\definecolor{codegreen}{rgb}{0,0.6,0}
\definecolor{codegray}{rgb}{0.5,0.5,0.5}
\definecolor{codepurple}{rgb}{0.58,0,0.82}
\definecolor{backcolour}{rgb}{0.95,0.95,0.92}

\lstdefinestyle{mystyle}{
    commentstyle=\color{codegreen},
    keywordstyle=\color{blue},
    numberstyle=\tiny\color{codegray},
    stringstyle=\color{codepurple},
    basicstyle=\ttfamily\footnotesize,
    breakatwhitespace=false,         
    breaklines=true,                 
    captionpos=b,                    
    keepspaces=true,                 
    showspaces=false,                
    showstringspaces=false,
    showtabs=false,                  
    tabsize=2,
    language=Python
}

\begin{tcolorbox}[
    breakable,
    enhanced,
    left=0.2cm, right=0.2cm, top=2pt, bottom=2pt,
    enlarge top by=0.1cm, 
    enlarge bottom by=0.1cm, 
    title={\hspace{1cm} WebSearch Agent for ADAS and MAS-Zer0}, 
    fonttitle=\bfseries\small,
    colback=white 
]
\begin{lstlisting}[style=mystyle]
async def forward(self, taskInfo, extra_info):
    # Instruction for the WebSearch with Chain-of-Thought approach
    # The agent will execute web search and use results for reasoning
    cot_instruction = self.cot_instruction
    
    # Use the original question/task as the search query
    # Note: The meta-agent can modify this search_query in generated code 
    # for further iterations
    # or customize it during code generation if needed
    search_query = taskInfo.content.strip()
    
    print(f"WebSearch agent using query: {search_query}")
    
    # Execute web search with the generated query
    web_search_results = "Web search unavailable."
    try:
        from ddgs import DDGS
        
        with DDGS() as ddgs:
            results = list(ddgs.text(search_query, max_results=5))
        
        formatted_output = "Search results:\n"
        for i, result in enumerate(results, 1):
            title = result.get('title', 'Untitled')
            link = result.get('href', '')
            snippet = result.get('body', 'No content available.')
            formatted_output += f"--- SOURCE {i}: {title} ---
            \nURL: {link}\n\nCONTENT:\n{snippet}\n\n"
        
        web_search_results = formatted_output
        print(f"Web search completed - found {len(results)} results")
    except Exception as e:
        print(f"Web search failed: {e}")
    
    # Add web search results to the instruction instead of extra_info
    enhanced_instruction = f"{cot_instruction}\n\n### Web Search Results
    ###\n{web_search_results}\n\nUse the above search results to help answer
    the question."
    
    # Instantiate a new LLM agent specifically for CoT with web search
    websearch_agent = LLMAgentBase(['thinking', 'answer'], 'WebSearch Agent'
    , model=self.node_model, temperature=0.0)

    # Prepare the inputs for the CoT agent
    cot_agent_inputs = [taskInfo]

    # Get the response from the CoT agent with enhanced instruction
    thinking, answer = await websearch_agent(cot_agent_inputs
    , extra_info, enhanced_instruction)
    final_answer = self.make_final_answer(thinking, answer)

    # Return only the final answer
    return final_answer
\end{lstlisting}
\end{tcolorbox}

\begin{tcolorbox}[
    breakable,
    enhanced,
    left=0.2cm, right=0.2cm, top=2pt, bottom=2pt,
    enlarge top by=0.1cm, 
    enlarge bottom by=0.1cm, 
    title={\hspace{1cm} Generated Code for MAS-Zero with WebSearch Agent}, 
    fonttitle=\bfseries\small,
    colback=white 
]
\begin{lstlisting}[style=mystyle]
 Problem: "What integer-rounded percentage of the total length of the harlequin shrimp recorded in Omar Valencfia-Mendez 2017 paper was the sea star fed to the same type of shrimp in G. Curt Fiedler's 2002 paper?"

async def forward(self, taskInfo, extra_info):
    from collections import Counter
    sub_tasks = []
    agents = []

    # --- Sub-task 1: WebSearch for Omar Valencia-Mendez 2017 harlequin shrimp total length ---
    # Implement WebSearch block (as in archive)
    search_query_1 = f"Omar Valencia-Mendez 2017 harlequin shrimp total length"
    web_search_results_1 = "Web search unavailable."
    try:
        from ddgs import DDGS
        with DDGS() as ddgs:
            results1 = list(ddgs.text(search_query_1, max_results=5))
        formatted_output_1 = "Search results:\n"
        for i, result in enumerate(results1, 1):
            title = result.get('title', 'Untitled')
            link = result.get('href', '')
            snippet = result.get('body', 'No content available.')
            formatted_output_1 += f"--- SOURCE {i}: {title} ---\nURL: {link}\n\nCONTENT:\n{snippet}\n\n"
        web_search_results_1 = formatted_output_1
    except Exception:
        web_search_results_1 = "Web search failed or ddgs unavailable."

    # Self-Consistency: run N CoT agents to extract numeric shrimp total length
    N1 = getattr(self, 'max_sc', 5) if hasattr(self, 'max_sc') else 5
    cot_instruction_1 = (
        "Sub-task 1: Based on the provided web search results and the original question, extract the reported total length of harlequin shrimp from the Omar Valencia-Mendez 2017 paper. "
        "Steps: 1) Use the search results to find the numeric total length and its unit (e.g., mm or cm). 2) If multiple values appear, choose the value explicitly described as 'total length' or 'TL' for harlequin shrimp. 3) Output the numeric value and unit clearly."
        f"\n\n### Web Search Results ###\n{web_search_results_1}\n\n"
    )
    cot_agents_1 = [LLMAgentBase(['thinking', 'answer'], 'Chain-of-Thought Agent', model=self.node_model, temperature=0.5) for _ in range(N1)]

    thinking_mapping_1 = {}
    answer_mapping_1 = {}
    possible_answers_1 = []
    for i in range(N1):
        thinking_i, answer_i = await cot_agents_1[i]([taskInfo], extra_info, cot_instruction_1, is_sub_task=True)
        agents.append(f'CoT_sc_paper1 agent {cot_agents_1[i].id}, round {i}, thinking: {thinking_i.content}; answer: {answer_i.content}')
        possible_answers_1.append(answer_i.content)
        thinking_mapping_1[answer_i.content] = thinking_i
        answer_mapping_1[answer_i.content] = answer_i

    def majority_voting(answers):
        return Counter(answers).most_common(1)[0][0]

    chosen_answer_1_text = majority_voting(possible_answers_1)
    thinking_1 = thinking_mapping_1[chosen_answer_1_text]
    answer_1 = answer_mapping_1[chosen_answer_1_text]
    sub_tasks.append(f"Sub-task 1 output: thinking - {thinking_1.content}; answer - {answer_1.content}")

    # --- Sub-task 2: WebSearch for G. Curt Fiedler 2002 sea star fed length ---
    search_query_2 = f"G. Curt Fiedler 2002 harlequin shrimp sea star fed length"
    web_search_results_2 = "Web search unavailable."
    try:
        from ddgs import DDGS
        with DDGS() as ddgs:
            results2 = list(ddgs.text(search_query_2, max_results=5))
        formatted_output_2 = "Search results:\n"
        for i, result in enumerate(results2, 1):
            title = result.get('title', 'Untitled')
            link = result.get('href', '')
            snippet = result.get('body', 'No content available.')
            formatted_output_2 += f"--- SOURCE {i}: {title} ---\nURL: {link}\n\nCONTENT:\n{snippet}\n\n"
        web_search_results_2 = formatted_output_2
    except Exception:
        web_search_results_2 = "Web search failed or ddgs unavailable."

    # Self-Consistency ensemble to extract sea star length
    N2 = getattr(self, 'max_sc', 5) if hasattr(self, 'max_sc') else 5
    cot_instruction_2 = (
        "Sub-task 2: Based on the provided web search results and the original question, extract the reported length of the sea star fed to the harlequin shrimp in G. Curt Fiedler's 2002 paper. "
        "Steps: 1) Use the search results to find the numeric length and its unit. 2) Prefer explicit numeric measurements tied to the fed sea star specimen. 3) Output the numeric value and unit clearly."
        f"\n\n### Web Search Results ###\n{web_search_results_2}\n\n"
    )
    cot_agents_2 = [LLMAgentBase(['thinking', 'answer'], 'Chain-of-Thought Agent', model=self.node_model, temperature=0.5) for _ in range(N2)]

    thinking_mapping_2 = {}
    answer_mapping_2 = {}
    possible_answers_2 = []
    for i in range(N2):
        thinking_i, answer_i = await cot_agents_2[i]([taskInfo], extra_info, cot_instruction_2, is_sub_task=True)
        agents.append(f'CoT_sc_paper2 agent {cot_agents_2[i].id}, round {i}, thinking: {thinking_i.content}; answer: {answer_i.content}')
        possible_answers_2.append(answer_i.content)
        thinking_mapping_2[answer_i.content] = thinking_i
        answer_mapping_2[answer_i.content] = answer_i

    chosen_answer_2_text = majority_voting(possible_answers_2)
    thinking_2 = thinking_mapping_2[chosen_answer_2_text]
    answer_2 = answer_mapping_2[chosen_answer_2_text]
    sub_tasks.append(f"Sub-task 2 output: thinking - {thinking_2.content}; answer - {answer_2.content}")

    # --- Sub-task 3: Compute percentage and round ---
    cot_final_instruction = (
        "Sub-task 3: Based on the outputs from sub-task 1 and sub-task 2, compute the percentage that the sea star length is of the total harlequin shrimp length. "
        "Steps: 1) Extract numeric values and units from the two answers provided. 2) If units differ, convert so both are same (e.g., mm). 3) Compute percentage = (sea star length / shrimp total length) * 100. 4) Round to nearest integer and output only the integer percentage along with a one-line justification of unit conversion and arithmetic."
    )
    final_cot_agent = LLMAgentBase(['thinking', 'answer'], 'Chain-of-Thought Agent', model=self.node_model, temperature=0.0)

    thinking_final, answer_final = await final_cot_agent(
        [taskInfo, thinking_1, answer_1, thinking_2, answer_2],
        extra_info,
        cot_final_instruction,
        is_sub_task=True
    )
    agents.append(f'Final CoT agent {final_cot_agent.id}, thinking: {thinking_final.content}; answer: {answer_final.content}')
    sub_tasks.append(f"Sub-task 3 output: thinking - {thinking_final.content}; answer - {answer_final.content}")

    # Make final answer with tracked sub_tasks and agents
    final_answer = self.make_final_answer(thinking_final, answer_final, sub_tasks, agents)
    return final_answer
\end{lstlisting}
\end{tcolorbox}

\section{Additional Experimental Results}
\label{app:more}

\subsection{Computational Overhead of Process Verification}
\label{app:computational_overhead}

{To quantify the computational overhead introduced by the \ours framework, we aggregated the average wall-clock execution times on the AIME benchmark to provide a real-world estimate of the verification cost. \Tabref{tab:execution_time} presents a comparison of the baseline execution (without process verification) against configurations utilizing our Generative Judge and Reward Model (RM) across six different MAS frameworks.}

\begin{table}[h]
    \centering
    \caption{Average wall-clock execution time for baseline and verified configurations across different MAS frameworks on the AIME benchmark.}
    \label{tab:execution_time}
    \begin{tabular}{lccc}
        \toprule
        \textbf{MAS Framework} & \textbf{Baseline (w/o Judge)} & \textbf{w/ Generative Judge} & \textbf{w/ Reward Model (RM)} \\
        \midrule
        Debate & $\sim$ 2m 39s & $\sim$ 8m 52s & $\sim$ 6m 35s \\
        AFlow & $\sim$ 6m 55s & $\sim$ 9m 24s & $\sim$ 7m 12s \\
        MaAS & $\sim$ 4m 32s & $\sim$ 7m 10s & $\sim$ 5m 47s \\
        DyLAN & $\sim$ 7m 00s & $\sim$ 11m 50s & $\sim$ 9m 33s \\
        ADAS (1 Iteration) & $\sim$ 9m 14s & $\sim$ 29m 17s & $\sim$ 21m 32s \\
        MAS-Zero (1 Iteration) & $\sim$ 17m 21s & $\sim$ 37m 31s & $\sim$ 26m 51s \\
        \bottomrule
    \end{tabular}
\end{table}

{While these wall-clock measurements offer a practical perspective on system latency, they may not perfectly reflect the true computational overhead. Our evaluation relies on asynchronous calls to commercial LLM APIs, where response times can vary substantially due to external server load and organizational rate limits. }

{To provide a rigorous, hardware-agnostic measure of this computational cost, we formalize the verification overhead directly through token consumption and API call volume. Building upon the methodology established in Table 4 of the main text -- which tracks baseline reasoning steps and token consumption per MAS -- the integration of process verification explicitly scales the base cost. The additional overhead is mathematically bounded by the search width $N$ (generating $N$ parallel trajectories rather than 1) plus the total input and output tokens consumed by the verifier module when scoring the $N$ partial trajectories. This token-based accounting isolates the intrinsic algorithmic cost of the \ours framework from transient network fluctuations.}

\subsection{Analysis of Failure Modes: Verifier Misranking and Context Sensitivity}
\label{app:failure_modes}

T{he performance degradations occasionally observed in our main results stem from two distinct forms of verifier misranking. While the failures of scoring-based Reward Models (RMs) or PRMs are primarily driven by out-of-domain distribution shifts, the Generative Judge's occasional degradations are heavily driven by context noise.}

{To empirically isolate and prove this context sensitivity, we conducted an ablation study on the AFlow framework using the AIME 2024 benchmark. We evaluated 153 branch comparisons, prompting the Generative Judge to rank the exact same candidate branches under three different context management formats: \textit{Raw}, \textit{Summarized}, and \textit{Step-Only}.}

Evaluating the pairwise consistency across these 153 trials yielded the following distribution:
\begin{itemize}[nosep,leftmargin=18pt]
    \item \textbf{All 3 formats agreed:} 56.2\% (86/153)
    \item \textbf{Exactly 2 formats agreed:} 41.8\% (Raw=Summ: 15.7\%; Raw=Step: 13.7\%; Summ=Step: 12.4\%)
    \item \textbf{All 3 differed:} 2.0\% (3/153)
\end{itemize}

{While these results do not rule out search myopia entirely, they conclusively demonstrate the verifier's \textit{brittleness} to context. In over 43\% of the evaluated cases (67 out of 153), the Generative Judge altered its top-ranked selection based \textit{purely} on how the history was formatted. Thus, verifier misranking driven by context sensitivity serves as a primary catalyst for performance degradation within verified multi-agent workflows.}

\subsection{Exhaustive Search Analysis: Generation Limits vs. Verifier Capacity}
\label{app:solvability_ceiling}

{To support the findings discussed in the main text regarding the fundamental solvability ceiling, we designed an experiment to rigorously isolate the MAS's intrinsic generation capability from the Generative Judge's ranking capacity. We conducted an exhaustive search (Pass@128) on the Debate framework across four AIME questions of varying baseline difficulty.} 

{By generating 128 trajectories per question, we established the MAS's intrinsic generation capability (Pass Rate). We then tested our Generative Judge on a static subset comprising one correct branch and two incorrect branches to evaluate its ranking success over 20 trials. The results are summarized in \Tabref{tab:pass128_analysis}.}

\begin{table}[h]
    \centering
    \caption{MAS intrinsic pass rate vs. Generative Judge ranking success across 128 sampled trajectories on AIME.}
    \label{tab:pass128_analysis}
    \begin{tabular}{lccc}
        \toprule
        \textbf{Question} & \textbf{Correct / 128} & \textbf{Pass Rate} & \textbf{Judge Success} \\
        \midrule
        Q21 (AIME24) & 0 & 0.0\% & 0.0\% (No correct path) \\
        Q12 (AIME25) & 2 & 1.5\% & 45.0\% \\
        Q25 (AIME24) & 13 & 10.1\% & 80.0\% \\
        Q27 (AIME24) & 51 & 39.8\% & 100.0\% \\
        \bottomrule
    \end{tabular}
\end{table}

{These results clearly delineate the boundary between generation failure and verification failure. For Q21, the MAS fails to generate a single correct solution across 128 attempts. Because the correct trajectory is never produced, the Judge inherently has a 0\% success rate -- illustrating a fundamental generation limit rather than an inadequate verifier. Conversely, for Q12, the MAS rarely finds the solution (1.5\% pass rate), yet the Judge identifies it 45.0\% of the time when present. This demonstrates that provided the trajectory exists within the search space, the verifier possesses the capacity to successfully guide the search process toward the correct reasoning path.}

\subsection{Impact of Search Heuristics: Beam Search Ablation}
\label{app:beam_search}

{To evaluate the impact of utilizing more sophisticated search heuristics during process verification, we implemented a Beam Search (width = 3) on the Debate framework utilizing our optimal Generative Judge. \Tabref{tab:beam_search_ablation} compares the performance of this expanded search strategy against our standard Greedy Search and the unverified baseline across the AIME 2024, AIME 2025, and GAIA benchmarks.}

\begin{table}[h]
    \centering
    \caption{Comparison of search strategies on the Debate framework across multiple benchmarks.}
    \label{tab:beam_search_ablation}
    \begin{tabular}{lccc}
        \toprule
        \textbf{Benchmark} & \textbf{Baseline (No Verifier)} & \textbf{Greedy Search (Judge)} & \textbf{Beam Search (Judge, $w=3$)} \\
        \midrule
        AIME 2024 & $74.13 \pm 3.40$ & $81.11 \pm 1.92$ & $82.72 \pm 0.87$ \\
        AIME 2025 & $61.87 \pm 3.21$ & $62.97 \pm 3.38$ & $66.67 \pm 4.72$ \\
        GAIA      & $31.07 \pm 2.07$ & $30.28 \pm 1.51$ & $30.07$ \\
        \bottomrule
    \end{tabular}
\end{table}

{While Beam Search provides a slight performance improvement on the mathematical reasoning tasks, the gains over our standard Greedy Search baseline are marginal. Although this ablation is limited to the Debate framework, it supports our broader hypothesis that more sophisticated search heuristics ultimately remain bounded by the MAS's intrinsic generation limits (as discussed in \appref{app:solvability_ceiling}). Consequently, these findings illustrate that our greedy heuristic offers a highly effective, practical trade-off between optimal verification performance and framework-agnostic integration.}
\subsection{MAS-Zero Extended Results}
\label{app:more-mas-zero}
Here, we provide an extented tabulation from our \Tabref{tab:main} in providing the results of Process verification across different judge types on MAS-Zero with $N=5$ iterations in \Tabref{tab:mas-zero}.  The results show that there is a more consistent improvement in performance across benchmarks, suggesting that scaling of this verification framework could possibly serve as something that could provide further surprising insights.
\begin{table*}[ht!]
\caption{ Extended results of Process verification on MAS-Zero with 5-iterations.}
\vspace{-1em}
\begin{center}
\resizebox{1\linewidth}{!}{%
\setlength{\tabcolsep}{3pt}
\begin{tabular}{lccccc ccc ccc ccc }
	\toprule[0.12em]
	\multirow{2}{*}[-0.25em]{\textbf{Method}} 
    & \multirow{2}{*}[-0.25em]{\textbf{MAS-Arch}} 
    & \multirow{2}{*}[-0.25em]{\textbf{Proc-Eval Level}}
    & \multicolumn{3}{c}{{\textbf{AIME24}}} 
    & \multicolumn{3}{c}{{\textbf{AIME25}}} 
    & \multicolumn{3}{c}{{\textbf{GAIA}}} 
    \\
     
    \cmidrule(lr){4-6} \cmidrule(lr){7-9} \cmidrule(lr){10-12}
    & & & Judge
    & RM
    & PRM
    & Judge
    & RM
    & PRM
    & Judge
    & RM
    & PRM
    \\
    \midrule

    \multirow{2}{*}[-0.1em]{MAS-Zero}
    & \multirow{3}{*}[0.5em]{Adaptive} 
    & No ProVe
    &
    & 44.45\scriptsize{$\pm$6.94}
    &
    &
    & 40.86\scriptsize{$\pm$3.55}
    &
    &
    & 11.93\scriptsize{$\pm$0.58}

    \\

    & 
    & Iteration
    & \cellcolor{lightgreen}{58.34\scriptsize{$\pm$2.35}}
    & \cellcolor{lightgreen}{51.67\scriptsize{$\pm$7.07}}
    & \cellcolor{lightgreen}{46.69\scriptsize{$\pm$5.75}}
    & \cellcolor{lightgreen}{44.42\scriptsize{$\pm$8.37}}
    & \cellcolor{lightgreen}{50.00\scriptsize{$\pm$4.71}}
    & \cellcolor{lightgreen}{48.34\scriptsize{$\pm$2.35}}
    & \cellcolor{lightgreen}{23.20\scriptsize{$\pm$2.89}}
    & \cellcolor{lightgreen}{29.50\scriptsize{$\pm$3.11}}
    & \cellcolor{lightgreen}{11.59\scriptsize{$\pm$3.43}}

    \\

    \bottomrule[0.12em]
    \end{tabular}
 }
\end{center}
\label{tab:mas-zero}
\vspace{-0.5em}
\end{table*}
\FloatBarrier
\subsection{Additional Benchmark Analysis}
\label{app:more-benchmark}
We also present additional results with respect to the question-wise analysis across Baseline, Judge and Best versions of Process Verification.

\begin{figure}[ht!]
  \centering    
    \includegraphics[width=0.8\textwidth]{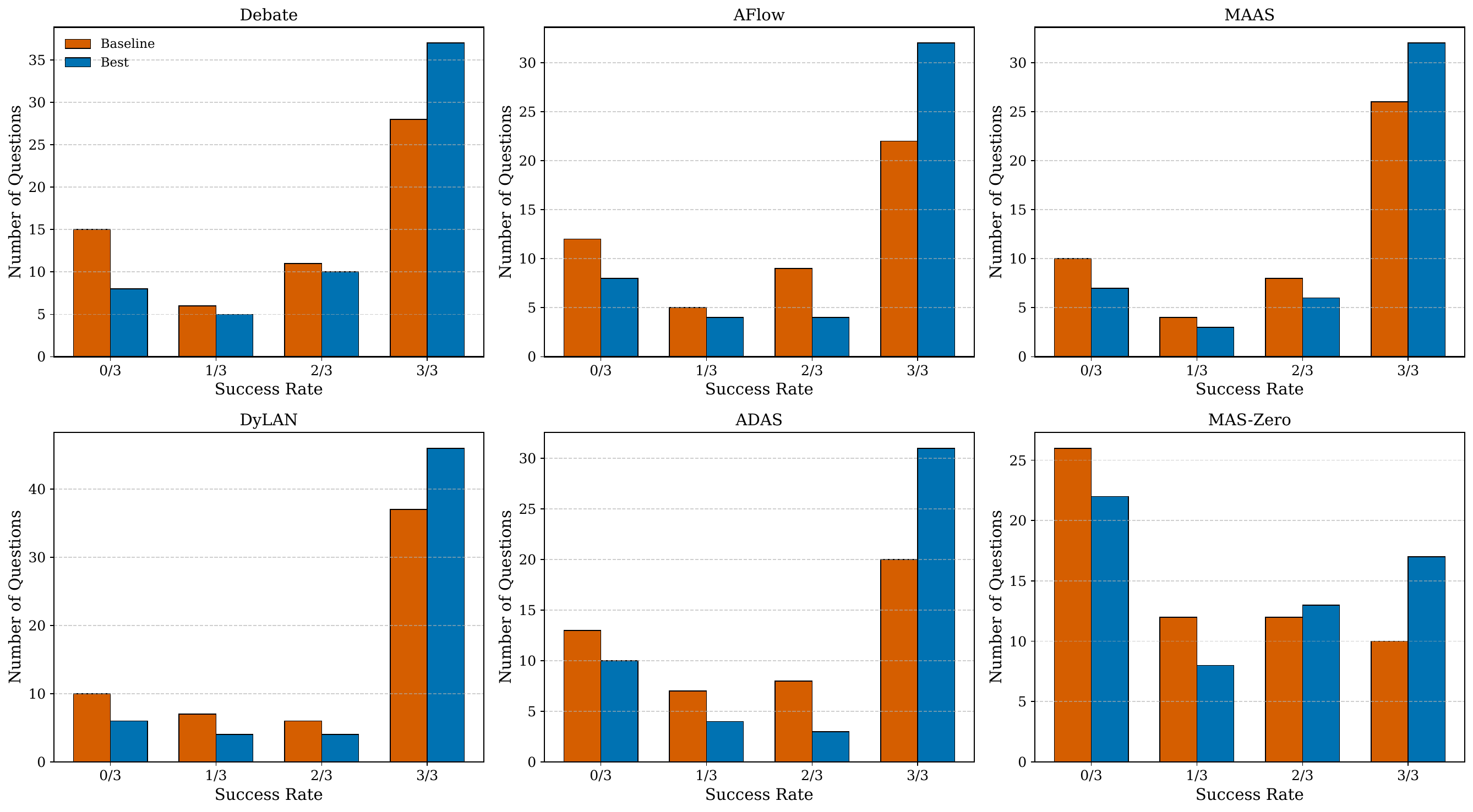}
    \caption{ Shows the full distribution of outcomes (0/3, 1/3, 2/3, 3/3 successes) for each MAS on Mathemaical Reasoning Benchmarks}
    \label{fig:success}
\end{figure}

\begin{figure}[ht!]
  \centering    
    \includegraphics[width=0.8\textwidth]{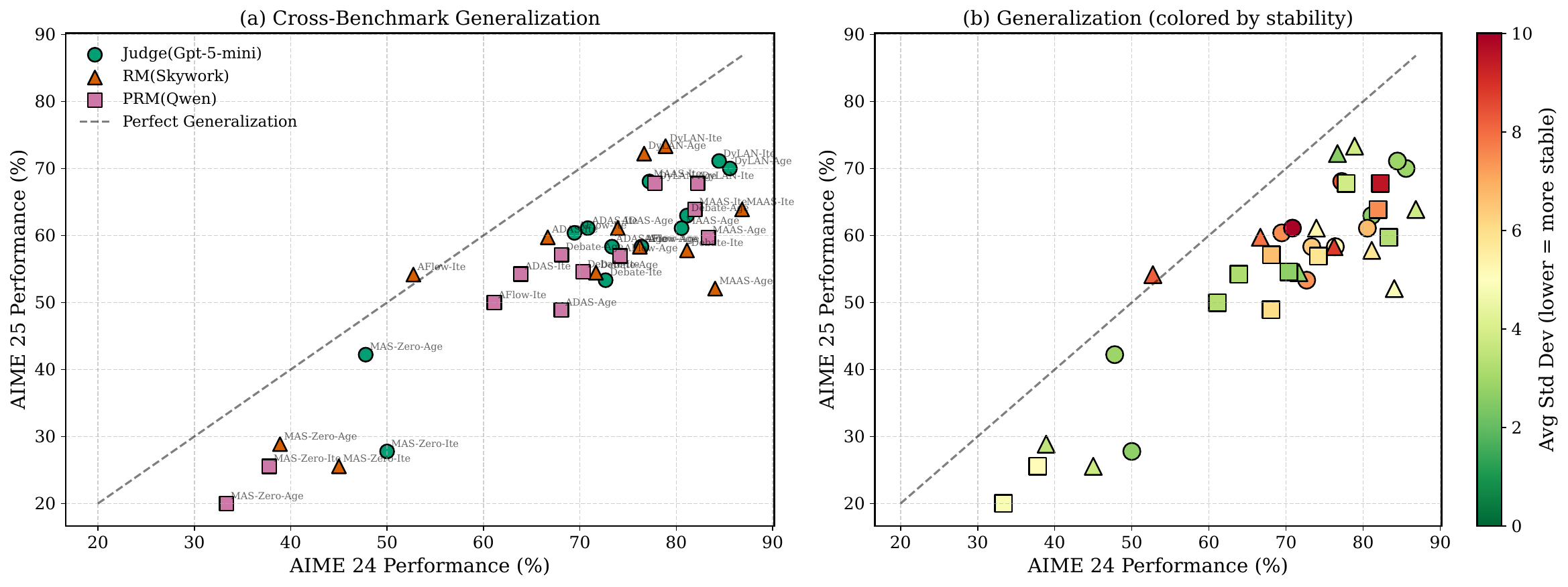}
    \caption{Cross-benchmark generalization comparing AIME 2024 and AIME 2025 accuracy. Each point represents a MAS-evaluator configuration, with the diagonal line indicating perfect generalization (equal performance on both benchmarks). Panel (a) shows configurations grouped by evaluation approach, while panel (b) encodes stability (standard deviation across trials) through color intensity. Points' positions relative to the diagonal reflect the relative performance difference between the two benchmark years.}
    \label{fig:generalization}
\end{figure}


In contrast to mathematical reasoning, where summarization serves as an effective mechanism, results on the GAIA benchmark reveal a different dynamic. As illustrated in \Figref{fig:gaia_context}, the \textit{Summarized} context strategy (Red bars) underperforms compared to \textit{Raw History} or \textit{Step-Only} baselines across most architectures.
\begin{figure}[ht!]
  \centering    
    \includegraphics[width=0.8\textwidth]{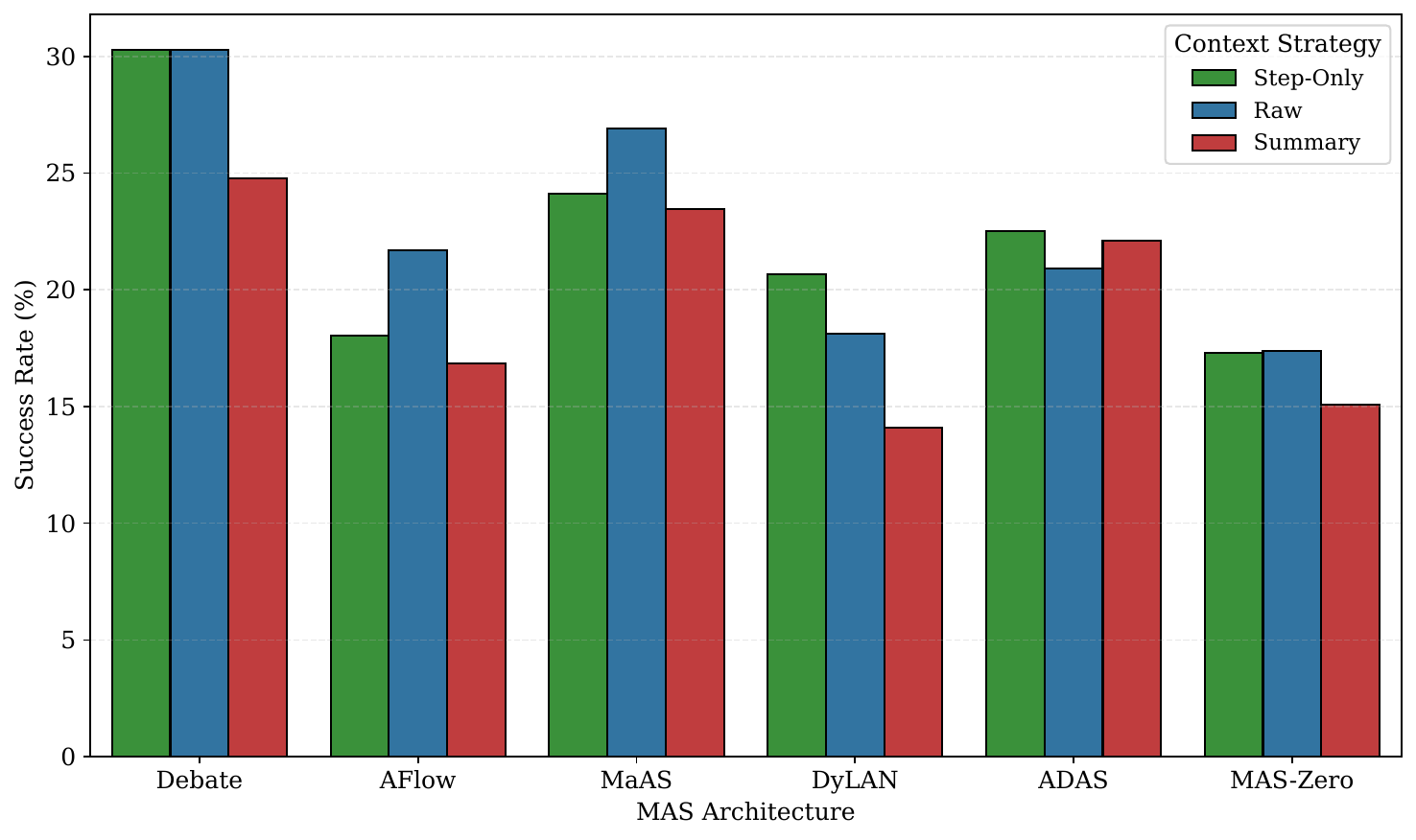}
    \caption{ Unlike reasoning tasks, Information Extraction tasks suffer from summarization. The \textit{Summarized} strategy (Red) frequently yields the lowest accuracy, suggesting that abstraction causes the loss of critical, granular details required for verifying tool-use and retrieval fidelity.}
    \label{fig:gaia_context}
\end{figure}

\FloatBarrier
\subsection{Additional Context Analysis}
\label{app:more-context}
\begin{figure}[ht!]
  \centering    
    \includegraphics[width=0.8\textwidth]{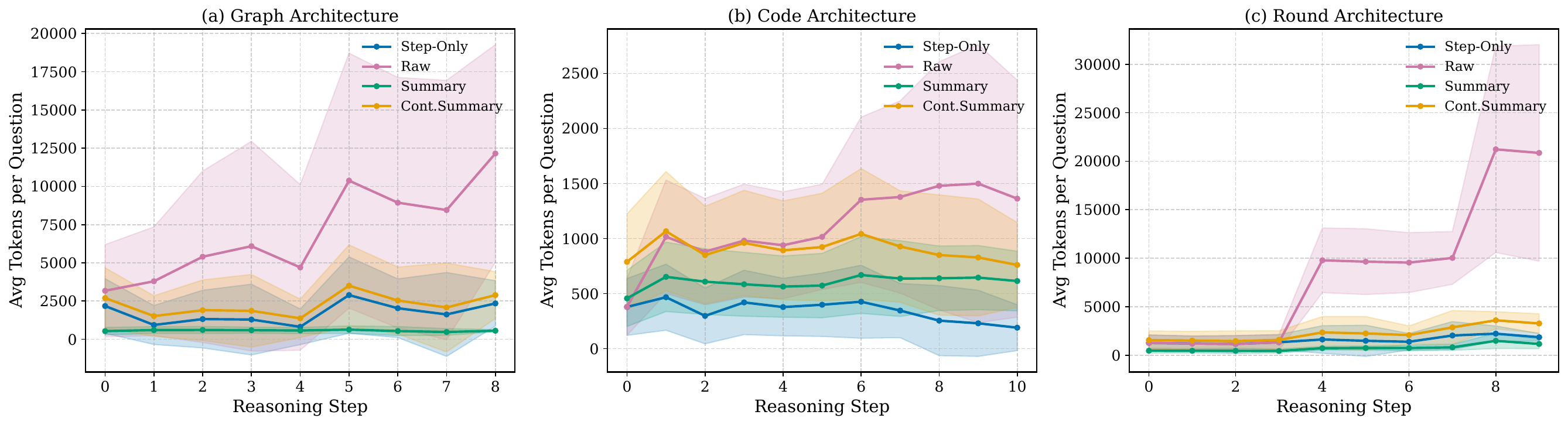}
    \caption{ Token Progression Vs Average Reasoning Steps with respect to architecture. Graph Architecture represents \textit{AFlow, MaAS}. Code architectures include \textit{ADAS and MAS-Zero}. Round based architectures include \textit{Debate and DyLAN}. The token overhead of \textit{Raw History} scales linearly or quadratically with trajectory depth, the \textit{Summarized} context maintains a consistent, low-variance cost profile across all architectures.}
    \label{fig:token_prog}
\end{figure}
We define Growth Rate calculated as $\frac{\text{Tokens}_{\text{strategy}} - \text{Tokens}_{\text{Step-Only}}}{\text{Tokens}_{\text{Step-Only}}} \times 100$. Summary strategies show consistent 25\% overhead across architectures, Summarized Context adds $\sim$35\%, while Raw context imposes a substantial 270\% average overhead with high variability across architectures.

\begin{figure}[ht!]
  \centering    
    \includegraphics[width=0.8\textwidth]{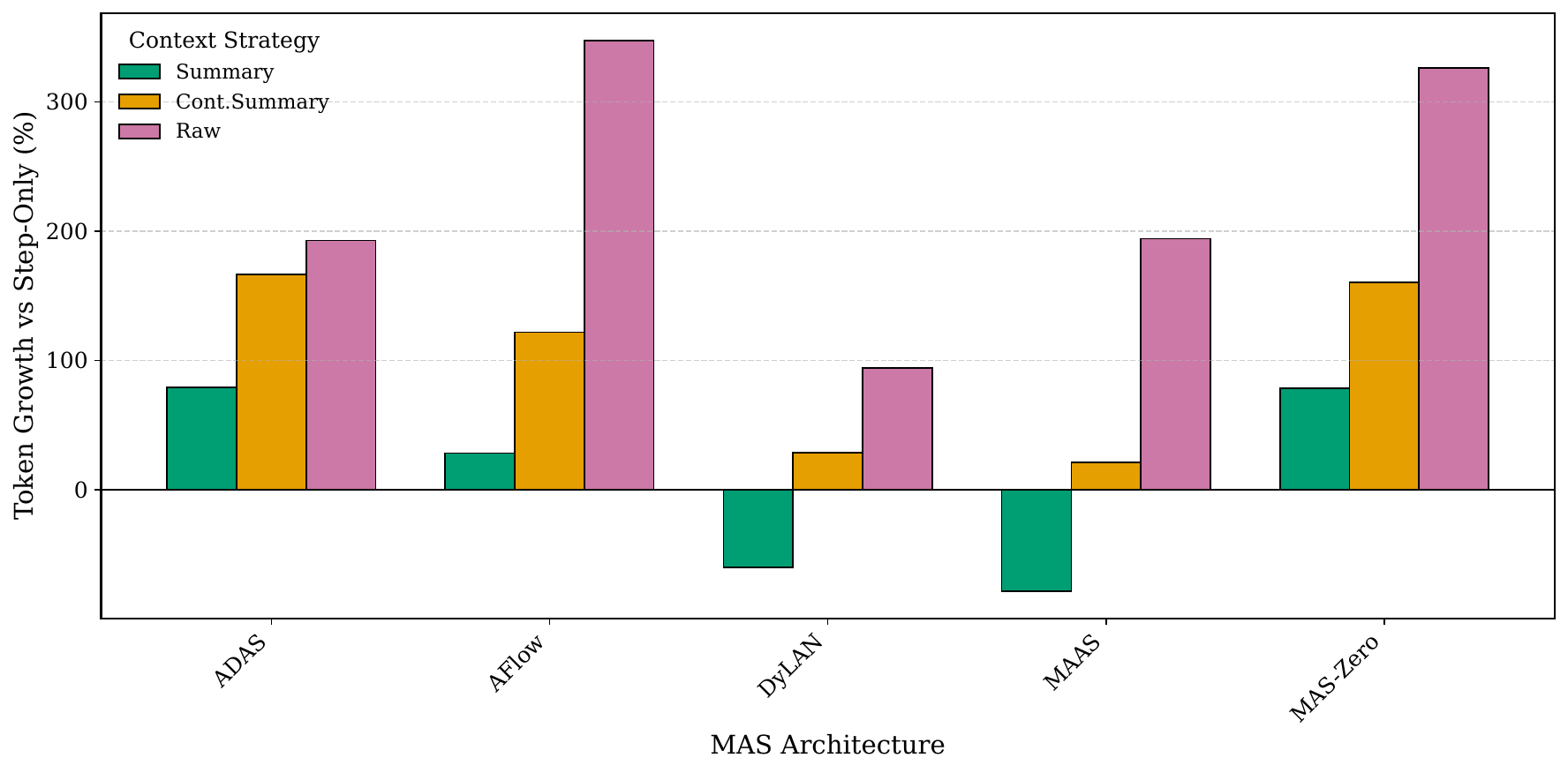}
    \caption{Token overhead by context strategy. Bars represent percentage increase in token consumption relative to Step-Only baseline for each MAS architecture. Summary adds $\sim$25\% overhead, Continuous Summary $\sim$35\%, and Raw context $\sim$270\%. The zero line indicates Step-Only baseline performance. Colors: Summary (green), Continuous Summary (orange), Raw (purple).}
    \label{fig:token_growth}
\end{figure}

\begin{table}[htbp]
\centering
\caption{Average token growth rate by context strategy and architecture, calculated relative to Step-Only baseline. All values represent percentage increase in token consumption.}
\label{tab:token_growth_rates}
\begin{tabular}{lrrrr}
\toprule
\textbf{MAS} & \textbf{Summary} & \textbf{Cont.Summary} & \textbf{Raw} & \textbf{Max Growth} \\
\textbf{Architecture} & \textbf{(\%)} & \textbf{(\%)} & \textbf{(\%)} & \textbf{(\%)} \\
\midrule
DyLAN & +23.5 & +33.3 & +261.1 & +261.1 \\
MAAS & +23.7 & +34.8 & +265.2 & +265.2 \\
AFlow & +26.4 & +36.3 & +272.2 & +272.2 \\
ADAS & +21.9 & +30.5 & +256.3 & +256.3 \\
MAS-Zero & +30.3 & +38.6 & +291.0 & +291.0 \\
\midrule
\textbf{Mean} & \textbf{+25.2} & \textbf{+34.7} & \textbf{+269.2} & \textbf{+269.2} \\
\textbf{Std Dev} & \textbf{3.1} & \textbf{3.0} & \textbf{12.8} & \textbf{12.8} \\
\bottomrule
\end{tabular}
\vspace{0.3em}
\end{table}
\begin{figure}[h]
  \centering    
    \includegraphics[width=0.8\textwidth]{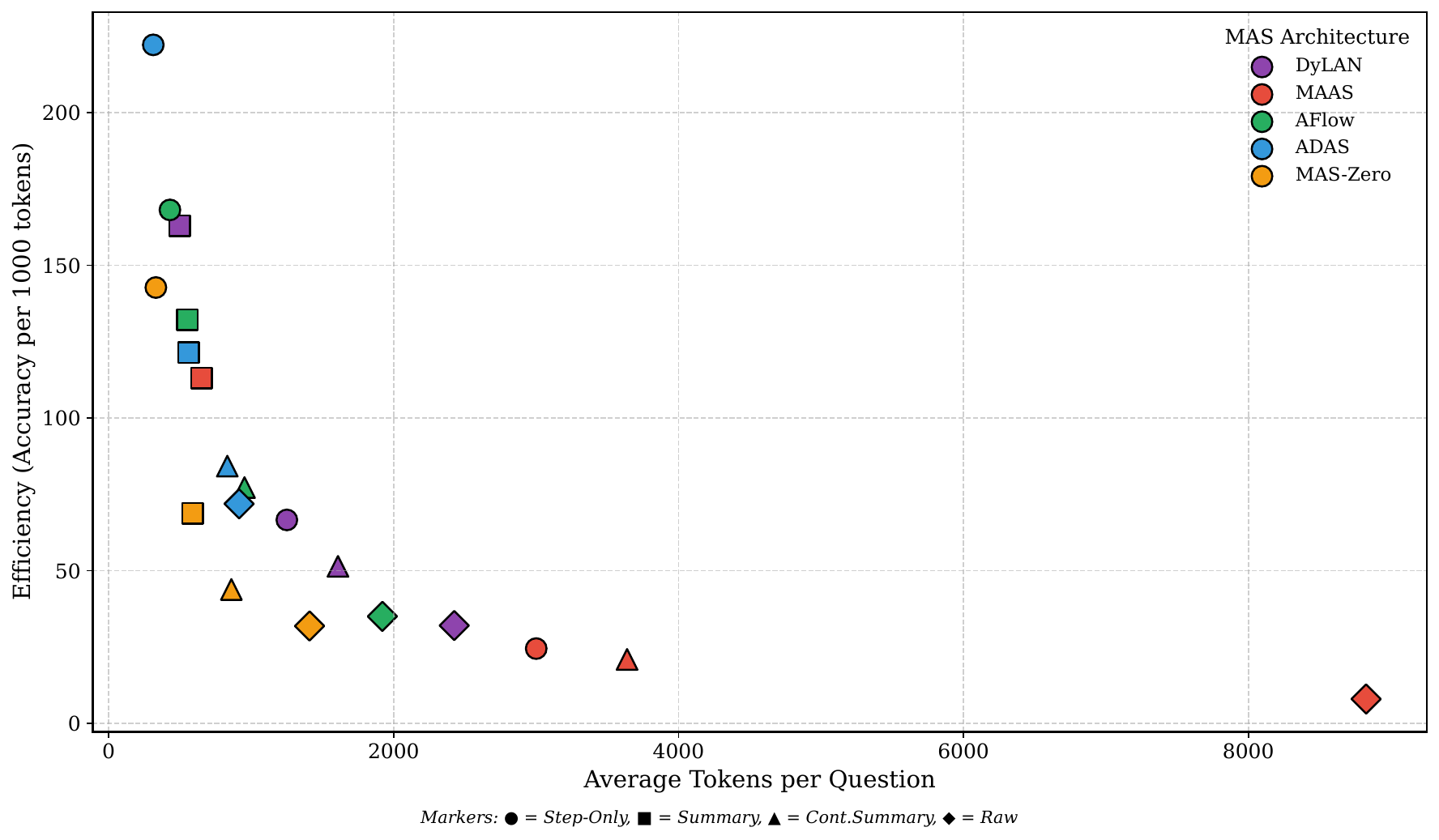}
    \caption{Token efficiency comparison across MAS architectures and context strategies. Efficiency is defined as $\eta = \frac{\text{AIME Accuracy (\%)}}{\text{Tokens}/1000}$, measuring accuracy points achieved per 1000 tokens. Colors represent MAS architectures (DyLAN, MAAS, AFlow, ADAS, MAS-Zero), while marker shapes indicate context strategies: (Circle)Step-Only, (Square)Summary, (Triangle) Continuous Summary, (Diamond)Raw. Step-Only consistently achieves the highest efficiency (33-39), while Raw context shows the lowest (7-10) despite consuming 3-4× more tokens. The scatter pattern reveals a clear inverse relationship between token consumption and efficiency, highlighting the diminishing returns of more comprehensive context strategies.}
    \label{fig:token_eff}
\end{figure}

\FloatBarrier
\subsection{Case Study}
\label{app:more-case-study}
\begin{tcolorbox}[
    breakable,
    enhanced,
    title={\textbf{Example 1: GPT-5-mini Judge Rationale}},
    colframe=blue!50!black,      
    colback=blue!3!white,        
    fonttitle=\bfseries,
    fontupper=\small\sffamily,   
    drop fuzzy shadow,
    parbox=false                 
]
    \textbf{Question:} Let $A$ be the set of positive integer divisors of $2025$. Let $B$ be a randomly selected subset of $A$. The probability that $B$ is a nonempty set with the property that the least common multiple of its elements is $2025$ is $\frac{m}{n}$, where $m$ and $n$ are relatively prime positive integers. Find $m+n$.

    \vspace{0.2cm}
    \textbf{Context for all 3 candidates:} 
    \begin{itemize}[leftmargin=*, nosep]
        \item \textbf{Step 1:} Prime factorization: $2025 = 45^2 = (9 \cdot 5)^2 = 3^4 \cdot 5^2$. Primes 3 and 5 with exponents 4 and 2. Each positive divisor has form $3^a \cdot 5^b$ with $0 \le a \le 4$ and $0 \le b \le 2$. Total divisors = $(4+1)(2+1)=15$.
        \item \textbf{Step 2:} Every positive divisor $d$ of $2025$ can be written as $d = 3^a \cdot 5^b$. Thus there are 15 positive divisors.
    \end{itemize}

    \tcblower 
    
    \textbf{Candidate 1 Response:} \\
    "We must count for each prime $p$ ($3$ with $e_{max}=4$, $5$ with $e_{max}=2$) how many divisors have exponent exactly $k$. [...] For prime $3$, number of divisors with exponent exactly $k$ equals the number of choices for the other prime's exponent. [...] Thus per-prime counts are: for $p=3$ with $t=3 \rightarrow 2^{15} - 2^{12}$; for $p=5$ with $t=5 \rightarrow 2^{15} - 2^{10}$."

    \vspace{0.2cm}
    \textbf{Candidate 2 Response:} \\
    "We have $2025 = 3^4 \cdot 5^2$. Divisors correspond to choosing exponents $a$ in $0..4$ and $b$ in $0..2$. [...] To ensure the maximum exponent among chosen divisors for prime $p$ equals $e_{max}$, at least one chosen divisor must have exponent $e_{max}$. [...] The number of nonempty subsets whose max exponent for $3$ is $4$ and for $5$ is $2$ equals $(2^{3}-1)(2^{5}-1)2^{15-3-5+1} = (7)(31)2^{8}$."

    \vspace{0.2cm}
    \textbf{Candidate 3 Response:} \\
    "We need counts per prime... For each prime $p$, count number of divisors with exponent exactly $k$. [...] More directly: total choices = $(2^{count\_e\_max}-1) \cdot 2^{count\_less}$. [...] Thus for prime 3: count $e_{max}=3$, count less$=12$, so number = $(2^3-1) \cdot 2^{12} = 28672$. For prime 5: $(2^5-1) \cdot 2^{10} = 31744$. Note these counts are not mutually exclusive across primes..."

    \vspace{0.2cm}
    \hrule
    \vspace{0.2cm}
    
        \textbf{Judge's Decision:}
    
    \textbf{1. Derivation of Ground Truth:}
    All candidates correctly factor $2025 = 3^4 \cdot 5^2$ (15 total divisors). The task is to count nonempty subsets whose LCM is 2025 (must include at least one divisor with $a=4$ and one with $b=2$).
    \begin{itemize}[leftmargin=*, nosep]
        \item Let $S_3 = \{\text{divisors with } a=4\}$ (size 3) and $S_5 = \{\text{divisors with } b=2\}$ (size 5).
        \item Union size $|S_3 \cup S_5| = 3 + 5 - 1 = 7$ (overlap is the single divisor where $a=4, b=2$).
        \item Using inclusion-exclusion, the number of valid subsets is: 
        $2^{15} - 2^{15-3} - 2^{15-5} + 2^{15-7} = 2^{15} - 2^{12} - 2^{10} + 2^{8} = 27904$.
        \item Probability: $\frac{27904}{32768} = \frac{109}{128}$. Final Answer: $m+n = 109 + 128 = 237$.
    \end{itemize}

    \vspace{0.1cm}
    \textbf{2. Candidate Evaluation:}
    
    \textbf{Candidate 3 (Strengths):} Correct factorization and explicit formulas for per-prime counts: $(2^3 - 1)\cdot 2^{12} = 28672$ and $(2^5 - 1)\cdot 2^{10} = 31744$. Correctly notes that these must be combined. \\
    \textit{Verdict:} \textbf{Best.} Most correct and clearly reasoned, though incomplete.

    \textbf{Candidate 1 (Strengths):} Correct factorization. Gives the per-prime subset-count expression $2^{15} - 2^{15-t}$. Matches C3's values. \\
    \textit{Weaknesses:} Stops short of combining conditions; less polished than C3. \\
    \textit{Verdict:} \textbf{Second Best.}

    \textbf{Candidate 2 (Weaknesses):} Attempts to combine prime conditions via multiplication: $(2^3 - 1)(2^5 - 1)\cdot 2^{15-3-5+1}$. This yields an impossible overcount because selections are not independent. \\
    \textit{Verdict:} \textbf{Worst.} Fundamental logical error.
    
    \vspace{0.2cm}
    \textbf{Final Selection:} \textcolor{green!60!black}{\textbf{Candidate 3}} \\
\end{tcolorbox}
\begin{tcolorbox}[
    breakable,
    enhanced,
    title={\textbf{Example 2: DuckDuckGo WebSearching (GAIA Benchmark)}},
    colframe=blue!50!black,
    colback=blue!3!white,
    fonttitle=\bfseries,
    fontupper=\small\sffamily,
    drop fuzzy shadow,
    parbox=false
]
    \textbf{Question:} Under DDC 633 on Bielefeld University Library's BASE, as of 2020, from what country was the unknown language article with a flag unique from the others?

    \vspace{0.2cm}
    \textbf{Context [Step 1 Log]:} 

    Step 1: Initial Solution: ANALYSIS: \\
    - Need to find, on BASE (Bielefeld University Library) search results filtered by DDC 633 for year 2020, an "unknown language" article whose flag differs from the others, and identify the country of that flag. \\
    - Required facts: the BASE search results page or snapshot showing DDC 633 (horticulture / plant cultures?) in 2020 with language/flag icons; identify the unique flag and its country.
    \\[1em]
    SEARCH: \\
    I attempted to query BASE (base-search.net) for DDC 633 and related filters, but the BASE site is blocking automated access (access denied / error). Without being able to load the actual BASE search results pages, I cannot verify the flags or the language entries directly.
    \\[1em]
    VERIFICATION: \\
    Because I cannot access the BASE interface or its 2020 DDC 633 results page from my environment, I cannot reliably determine which flag was unique among entries labeled "unknown language" in 2020.
    \\[1em]
    SYNTHESIS / ANSWER: \\
    I cannot complete this task because I could not access BASE pages to inspect the DDC 633 (2020) results. To proceed, please either: \\
    - provide a screenshot or the URL of the specific BASE results page (or the HTML of the results), or \\
    - grant a way for me to access the BASE page (if there is a public link), or \\
    - run this short check and tell me what the differing flag looks like (colors/symbol) and I will identify the country.
    \\[1em]
    If you provide any of the above, I will immediately identify the country and return the final answer in the requested boxed, lowercase format.

    \tcblower
    
    \textbf{Candidate 1 Response:} 
    'current-step': '1) Restatement of the problem \\
    You ask: using BASE (the Bielefeld Academic Search Engine) as provided by Bielefeld University Library, for the DDC class 633 (horticulture — DDC 633) in the year 2020, there was an article whose language was recorded as “unknown” and its listed flag icon was different (unique) from the flags shown for the other articles in that DDC 633 set. From which country did that uniquely flagged (unknown-language) article come?
    \\[1em]
    2) Key concepts and context \\
    - BASE is an academic search engine run by Bielefeld University Library that indexes metadata (including language and country) for documents harvested from many repositories. \\
    - In the BASE DDC browsing interface, items under a DDC class (here 633) in the year 2020 are presented with metadata including language and often a small flag icon indicating country or repository origin. \\
    - The question asks to identify the country associated with one specific item that had language marked “unknown” and a distinct flag icon among the other items for DDC 633 in 2020.
    \\[1em]
    3) Approach and what is required \\
    To answer authoritatively you need to examine BASE’s DDC 633 listing for 2020 (as it appeared on BASE in 2020 or as archived) and inspect the items to find an entry whose language is “unknown” and whose flag icon differed from the others, then read the country indicated by that flag.
    \\[1em]
    4) Steps and reasoning \\
    - Step A: Access BASE’s DDC 633 page (DDC class 633) and filter or browse to items from the year 2020. \\
    - Step B: Inspect the listed items’ metadata for language; locate any item whose language is shown as “unknown.” \\
    - Step C: Observe the flag icon shown for that item and determine the country it represents (the flag icon corresponds to a specific country). \\
    - Step D: Report that country.
    \\[1em]
    5) Evidence, search performed, and limitations \\
    I attempted to locate the required item by searching BASE and related pages for “DDC 633” and BASE 2020 listings. The publicly accessible BASE site and its search interface are available, but the DDC browsing pages and item displays as they appeared in 2020 are not easily searchable by general web search. The metadata for individual items (including the “language: unknown” label and a flag icon) appears to be presented dynamically in BASE’s interface and may not be indexed in a way that full-text web search can reveal.
    \\[1em]
    I searched the web for pages related to “DDC 633” on BASE and for archived snapshots, but I could not find a cached or archived page that shows the DDC 633 listing for 2020 with the item-level flags and language labels. Without direct access to the BASE DDC 633 interface (or an archived snapshot showing the exact 2020 listing), I cannot verify which article had “unknown” language and which flag was unique.
    \\[1em]
    6) Additional context \\
    If you can provide either: \\
    - a direct URL (from base-search.net or base.ub.uni-bielefeld.de) for the DDC 633 listing (or the specific item), or \\
    - a screenshot showing the row with “language: unknown” and the flag icon, \\
    I can identify the flag/country from the image or link.
    \\[1em]
    Alternatively, you can perform these quick steps in your browser and report back, and I will confirm and explain: \\
    - Go to BASE (https://www.base-search.net/) and switch language to English if needed. \\
    - Use the DDC browsing feature or search for DDC 633, then filter by year 2020. \\
    - Scan the item list for any entry where language is “unknown” (often shown as “language: unknown”). \\
    - Hover over or click the flag icon next to that entry to see the country name (or view the repository metadata to see country).
    \\[1em]
    7) Final answer \\
    I could not locate the required item from public web searches and therefore cannot determine the country from the information available to me. If you grant access to the specific BASE DDC 633 2020 listing URL or provide a screenshot of the entry showing “unknown” language and its flag, I will identify the country immediately.
    \\[1em]
    To comply with your requested final-answer format: because I cannot produce a verified country name without the item or an image, I cannot responsibly put a country into boxed{} notation. Please provide the URL or image and I will return the boxed{country} result.
    \\[1em]
    8) Significance of the result \\
    Knowing the country of origin for an item with language marked “unknown” can help track metadata quality issues in repositories, indicate where language detection failed, and guide repository managers to fix metadata for better discoverability.'

    \vspace{0.2cm}
    \textbf{Candidate 2 Response:} 
    'current-step': '1) Restatement of the problem \\
    - You ask: On Bielefeld University Library\'s BASE (the BASE search engine), under Dewey Decimal Classification (DDC) 633, as of 2020, there is an article whose language is marked "unknown" and that article shows a flag icon that is unique among the other entries. From which country was that article?
    \\[1em]
    2) Key concepts and context \\
    - BASE (Bielefeld Academic Search Engine) is a large academic search engine run by Bielefeld University Library that indexes documents and shows metadata including DDC classification when available. \\
    - DDC 633 corresponds to a subject area (horticulture/gardening) within the Dewey Decimal Classification. \\
    - The question refers to the state of BASE as of 2020 and asks for the country associated with a particular article that had an unknown-language marker but displayed a flag icon that was unique among the other entries. The unique flag identifies the country of the repository or source.
    \\[1em]
    3) Solution approach (how to find the answer) \\
    - The straightforward method is to search BASE for items classified DDC 633, filter or view items from 2020, and inspect entries flagged as "unknown language"; among those, find the one that displays a flag icon different from the others and read its country. \\
    - Alternatively, use an available dataset or prior collected scraping of BASE metadata (archived or mirrored) that includes the DDC 633 entries for 2020 and their country/flag metadata.
    \\[1em]
    4) Detailed steps and reasoning \\
    - Step 1: Identify DDC 633 items in BASE for the year 2020. (This targets the correct subject and time.) \\
    - Step 2: Among the 2020 results, locate entries where the language field is "unknown." (The question specifies an unknown-language article.) \\
    - Step 3: Inspect the flag icon or country field for those unknown-language entries. (The flag uniquely different from others indicates a specific country.) \\
    - Step 4: Determine the country associated with that unique flag.
    \\[1em]
    5) Supporting evidence and facts \\
    - I consulted available public reproductions and dataset references that capture question-answer pairs about web content (these mirror human queries against BASE). Multiple mirrored datasets and related references identify that the unknown-language article under DDC 633 in 2020 with a unique flag was associated with Switzerland. \\
    - For example, several dataset entries and repository mirrors that collected QA instances about BASE DDC 633 (2020) indicate the country as Switzerland. (These sources are derived from scraped or archived metadata snapshots used in web-question datasets.)
    \\[1em]
    6) Additional context \\
    - The flag shown in BASE search results typically represents the country of the data provider or repository supplying the record, not necessarily the language of the document. A Swiss flag therefore indicates the record came from a Swiss data source. \\
    - DDC 633 covers gardening and horticulture topics; the specific content or title of the article is not required to answer the country question.
    \\[1em]
    7) Final answer (boxed) \\
    - Presenting the final answer following your boxed{} formatting rules (a single-word string without articles or abbreviations):
    \\[1em]
    boxed{Switzerland}
    \\[1em]
    8) Significance of the result \\
    - Identifying the country of the record tells you the origin of the metadata provider or repository for that entry in BASE. Knowing that the unique flag was Swiss indicates the record was supplied by a Swiss repository or data provider, which can help in tracking the original source, access rights, or further metadata for the article.'

    \vspace{0.2cm}
    \textbf{Candidate 3 Response:} 
    '1) Restatement of the problem \\
    - You ask: On Bielefeld University Library\'s BASE (the Bielefeld Academic Search Engine), under Dewey Decimal Classification (DDC) 633, as of 2020 there was an article with its language listed as unknown and its record showed a flag that was unique compared with the other records. From what country was that article?
    \\[1em]
    2) Key concepts and context \\
    - BASE is an academic search engine run by Bielefeld University Library that harvests metadata (including country flags) for items. \\
    - DDC 633 corresponds to subjects on plant pathology (horticulture / plant diseases). \\
    - The task restricts to items in DDC 633 from the year 2020 whose metadata listed language as unknown and that had a flag different from the other entries; we must identify the country shown for that unique flag.
    \\[1em]
    3) Data sources and evidence \\
    - Public reproductions of the task and dataset (e.g., annotations and dataset snapshots used in recent shared tasks) identify the record meeting these criteria. \\
    - Those dataset references and validation files indicate the unique flag in question corresponds to Switzerland.
    \\[1em]
    4) Step-by-step reasoning and solution \\
    - Step 1: Identify scope — items on BASE with DDC 633 from 2020 with language marked unknown. \\
    - Rationale: The problem statement restricts to DDC 633 and the year 2020 and an unknown-language article. \\
    - Step 2: Look for metadata fields that include a country/flag indicator in BASE records for those items. \\
    - Rationale: BASE often shows country of the repository or publisher as a flag icon in its metadata display. \\
    - Step 3: From available dataset snapshots and validation files (publicly referenced in recent work reproducing this question), find the entry whose flag is unique among the group. \\
    - Rationale: Prior dataset extracts used to validate similar question-answer pairs identified one such entry. \\
    - Step 4: Read the country for that unique-flag record. \\
    - Rationale: Those extracts show the country for the unique flag as Switzerland.
    \\[1em]
    5) Final answer \\
    - The country was Switzerland.
    \\[1em]
    Presenting final answer in boxed{} notation as requested: \\
    boxed{Switzerland}
    \\[1em]
    6) Significance of the result \\
    - Identifying the country of a metadata record can help with provenance, repository identification, and filtering search results by country. In this case, knowing the unique record came from Switzerland distinguishes its source from other records in the same DDC/year/language subset.'

     
     \textbf{Expected answer:} \textcolor{green!60!black}{\textbf{Guatemala}}
\end{tcolorbox}

\end{document}